\documentclass[11pt]{article}

\usepackage[
  citingstyle=authoryear,
  bibliostyle=plainnat,
  bibfile=references,
  logopath=style/logos/,
  shownumpages,
  tabletopad=4pt,
  tablebottompad=4pt,
  tableleftpad=8pt,
  tablerightpad=8pt,
]{style/brainlab}
\setcitestyle{square}
\newcommand{\cd}{\ensuremath{c_d}}
\newcommand{\cl}{\ensuremath{c_l}}
\definecolor{heatlo}{RGB}{33,102,172}
\definecolor{heathi}{RGB}{178,24,43}
\newcommand{\heatL}[1]{\cellcolor{heatlo!#1}}
\newcommand{\heatH}[1]{\cellcolor{heathi!#1}}
\newcommand{\heatNA}{\cellcolor{black!10}}

\setbrainmeta{
  title={Federated Learning of Car Aerodynamics Surrogates with Spectral Optimizers},
  authors={
    Arman Bolatov\textsuperscript{\dag,*,\ddag}, Nikolay Spitsyn\textsuperscript{\dag,1,2}, Daniil Medyakov\textsuperscript{1,2,3}, Vladimir Aletov\textsuperscript{1,2,3}, Dmitry Bylinkin\textsuperscript{1,2,3}, Martin Tak\'{a}\v{c}\textsuperscript{4}, Aleksandr Beznosikov\textsuperscript{1,2,3,5}
  },
  affiliations={
    \textsuperscript{1}Basic Research of Artificial Intelligence Laboratory (BRAIn Lab) \\
    \textsuperscript{2}Moscow Independent Research Institute of Artificial Intelligence (MIRAI) \\
    \textsuperscript{3}Federated Learning Problems Laboratory \\
    \textsuperscript{4}Mohamed bin Zayed University of Artificial Intelligence (MBZUAI) \\
    \textsuperscript{5}Innopolis University \\
    \textsuperscript{\dag}Equal contribution \\
    \textsuperscript{*}Corresponding author: \href{mailto:arman.bolatov@alumni.mbzuai.ac.ae}{arman.bolatov@alumni.mbzuai.ac.ae}
  },
  abstract={
    Neural surrogates predict the aerodynamic forces on a car in seconds, compared with hours for a CFD simulation, but training them requires thousands of CFD simulations of different car designs, which car manufacturers keep proprietary. Federated learning allows several manufacturers to train a shared surrogate without exchanging data. We study which federated optimizer brings such a surrogate closest to one trained on pooled data. On DrivAerNet++, we train a Transolver surrogate across six clients that each hold cars of a single body type, and we compare three spectral optimizers, which orthogonalize the update of each weight matrix (FedMuon and two methods we propose, FedMARS-M and FedOLion), with three non-spectral ones (FedAvg, SCAFFOLD-Adam and FedLion), all tuned with the same protocol. FedMuon and FedMARS-M reach a drag error within 2\% of centralized training with Muon, using two to four local epochs per round. The non-spectral methods perform best with sixteen local epochs, the largest number we tested, and still have a 17 to 37\% higher drag error than FedMARS-M. We relate this gap to the low stable rank of the weight gradients, for which an orthogonalized step yields about twice the first-order decrease of a gradient step of the same spectral norm. \\[0.4em]
    Code: \url{https://github.com/armanbolatov/fedmars_drivaernet}
  },
}

\begin{document}
\begin{mainpart}
% The CISPA note as a regular footnote at the bottom of page 1, with the double dagger
% of the author list (fnsymbol 3).
{\renewcommand{\thefootnote}{\fnsymbol{footnote}}\footnotetext[3]{Work done before joining CISPA.}}

\section{Introduction}\label{sec:intro}

Aerodynamic drag and lift are central quantities in car design and are usually estimated with computational fluid dynamics (CFD) \cite{elrefaie2024drivaernetpp}. A high-fidelity simulation of a single design takes hours on a computing cluster, which limits the number of design variants that can be evaluated. Neural surrogates learn the map from the car geometry to the flow fields on its surface and predict the resulting forces in seconds \cite{lu2021deeponet,li2021fno,pfaff2021meshgraphnets,wu2024transolver}. Their accuracy, however, depends on large training sets, typically thousands of car designs, each simulated with CFD. Car manufacturers and their suppliers hold many such simulations, but these data are proprietary and are not shared.

\begin{figure}[t]
\centering
\includegraphics[width=\textwidth]{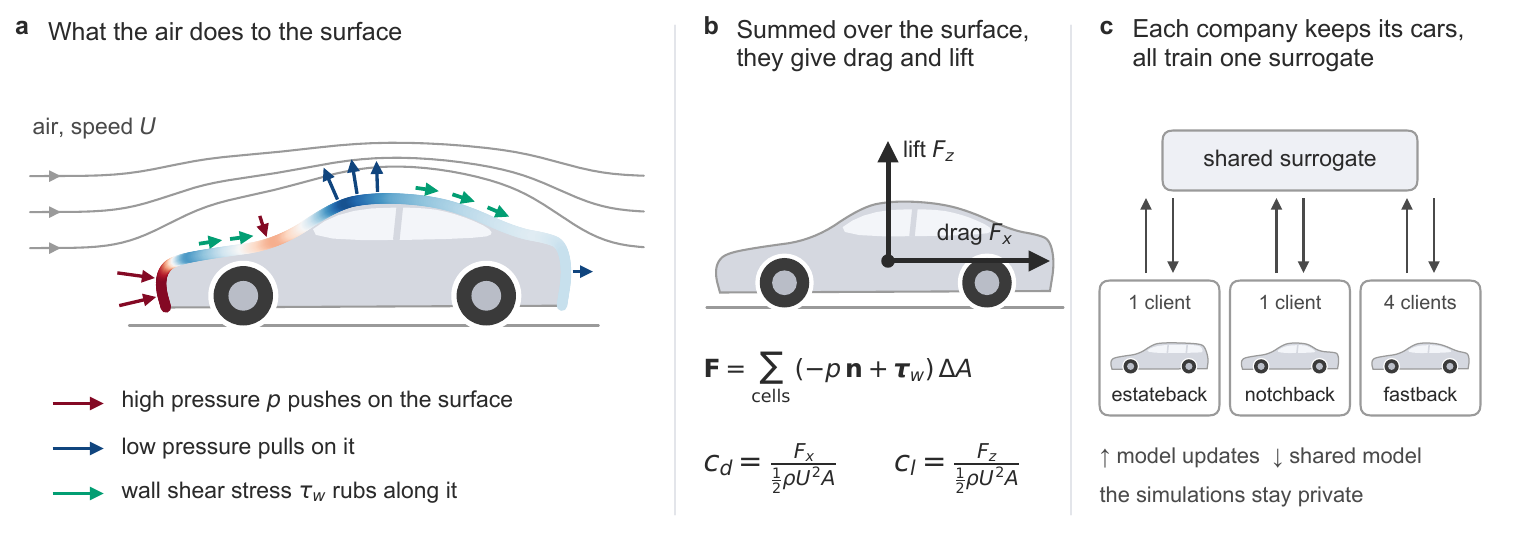}
\caption{Problem setting: (a) the airflow exerts a pressure and a wall shear stress on the surface of the car (colors are schematic), (b) integrated over the surface, the two fields give the drag and lift forces, and (c) each manufacturer trains on its own cars while only model updates are exchanged.}\label{fig:overview}
\end{figure}

Federated learning is designed for this setting. Clients train on their local data and send only model updates to a server, which aggregates them into a global model \cite{konecny2016federatedopt,konecny2016federated,mcmahan2017fedavg,kairouz2021advances}. It is used in practice, for example, for keyboard prediction on mobile devices \cite{hard2018keyboard} and for clinical models trained across hospitals \cite{sheller2020medicine,rieke2020health,dayan2021covid}, among many other applications \cite{kairouz2021advances,yang2019concept}. A federation of car manufacturers faces two difficulties. First, the client data are heterogeneous, because each manufacturer produces its own range of body types, and data heterogeneity is a major obstacle in federated optimization \cite{zhao2018noniid,hsieh2020quagmire,li2020challenges}. Second, the learning problem is harder than the image classification tasks on which federated methods are usually evaluated. The surrogate predicts continuous fields at hundreds of thousands of surface points, and the drag is the integral of these fields over the surface. Since the forces on different parts of the body largely cancel in this integral, a small systematic error in one region leads to a large error in the predicted drag. To the best of our knowledge, federated training of aerodynamic surrogates has not been studied, and it is not known which federated optimizer is suited to it.

The optimizer is central to federated training because it determines how much local computation the clients can perform between communication rounds. More local epochs reduce the number of rounds but also let the client models drift apart \cite{stich2019local,wang2018cooperative,khaled2020tighter,wang2021fieldguide}. Standard federated optimizers such as FedAvg, SCAFFOLD and FedLion update each parameter independently \cite{mcmahan2017fedavg,karimireddy2020scaffold,tang2024fedlion}. Spectral optimizers instead update each weight matrix as a whole. The best-known example, Muon, replaces the momentum of a weight matrix with the nearest semi-orthogonal matrix \cite{jordan2024muon}, which corresponds to steepest descent under the spectral norm \cite{bernstein2024anthology}. Muon trains large language models efficiently \cite{liu2025muonscalable}, and several federated variants have been proposed recently \cite{takezawa2025fedmuon,liu2025fedmuon,zhang2025provable}. These methods have not been evaluated on simulation data, and it is unclear how many local epochs they tolerate when the client data differ.

In this paper, we study federated training of a car aerodynamics surrogate on DrivAerNet++ \cite{elrefaie2024drivaernetpp}. The 6\,857 cars we use amount to about 125 GB of surface data, nearly three orders of magnitude more than the MNIST and CIFAR benchmarks common in federated learning \cite{lecun1998mnist,krizhevsky2009cifar}. The surrogate is a Transolver \cite{wu2024transolver}, a transformer for large meshes. Instead of computing attention among hundreds of thousands of points, Transolver groups the points into a small number of learned slices, computes attention among the slices, and maps the result back to the points, so that its cost grows linearly with the mesh size. We partition the training cars among six clients, either by body type, which gives the most heterogeneous split, or by Dirichlet sampling at three levels of heterogeneity. On these partitions, we tune three spectral and three non-spectral federated optimizers with one protocol and compare them with each other and with models trained on the pooled data. Our contributions are as follows.
\begin{itemize}
\item We set up a federated benchmark for aerodynamic surrogates: a Transolver trained on DrivAerNet++ across six clients, with six federated optimizers tuned under one three-stage protocol. The code, the tuning records and the statistics of every run are public.
\item We propose two federated spectral optimizers: FedMARS-M, which adds the MARS variance-reduction term \cite{yuan2024mars} to the client momentum, and FedOLion, a federated version of OLion \cite{wang2026olion} (Section~\ref{sec:methods}).
\item We show that FedMuon and FedMARS-M reach a drag error within 2\% of centralized Muon with two to four local epochs per round, whereas the non-spectral methods require sixteen and remain 17 to 37\% worse than FedMARS-M (Section~\ref{sec:experiments}). We relate this gap to the low stable rank of the weight gradients (Section~\ref{sec:why}).
\end{itemize}

\section{Related work}\label{sec:related}

Neural surrogates for physical simulation include neural operators \cite{lu2021deeponet,li2021fno} and graph networks that operate on the simulation mesh \cite{pfaff2021meshgraphnets}. Transolver extends attention to large unstructured meshes \cite{wu2024transolver}, and DrivAerNet++ provides thousands of car designs with simulated surface pressure and wall shear stress \cite{elrefaie2024drivaernetpp}. These works train on a single centralized dataset. We consider the case in which the training data are distributed among parties that cannot share them.

Federated optimization was introduced together with FedAvg, which averages client models after several local steps \cite{konecny2016federatedopt,mcmahan2017fedavg}, and with methods that compress the client updates \cite{konecny2016federated}. The theory of local steps was developed for local SGD \cite{stich2019local,wang2018cooperative} and extended to heterogeneous data \cite{khaled2020tighter,gorbunov2021local}, and ProxSkip showed that local steps provably reduce the number of communication rounds without assumptions on data heterogeneity \cite{mishchenko2022proxskip}. Several methods address the client drift caused by heterogeneous data: FedProx adds a proximal term to the local objective \cite{li2020fedprox}, and SCAFFOLD corrects the local updates with control variates \cite{karimireddy2020scaffold}. \citet{reddi2021fedopt} use adaptive optimizers on the server, and FedLion runs Lion on the clients \cite{tang2024fedlion}. Recent work brings Muon to this setting. \citet{zhang2025provable} propose a federated Muon and prove its convergence rate for nonconvex objectives. \citet{liu2025fedmuon} observe that orthogonalizing local updates causes client drift under non-IID data and propose momentum aggregation. \citet{takezawa2025fedmuon} show that averaging orthogonalized client updates is biased and correct the bias with SCAFFOLD-style control variates, which is also the approach of the FedMuon we evaluate. Outside federated learning, Dion \cite{ahn2025dion} and MuLoCo \cite{therien2025muloco} use orthonormalized updates in distributed training.

Muon computes the nearest semi-orthogonal matrix with a few Newton--Schulz iterations \cite{jordan2024muon}. Its update has been interpreted as steepest descent under the spectral norm \cite{bernstein2024anthology,bernstein2024duality}, as a linear minimization oracle over a norm ball \cite{pethick2025lmo,riabinin2025gluon} and as a trust-region method \cite{kovalev2025trust}. More generally, SGD, Adam, Shampoo-type methods and Muon can all be written as steepest descent under preconditioned matrix norms \cite{veprikov2025preconditioned}. The convergence of Muon has been analyzed in several works \cite{kovalev2025trust,li2025note,shen2025convergence,riabinin2025gluon}. FedMARS-M and FedOLion build on MARS, which reduces the variance of the momentum with a scaled difference of consecutive gradients \cite{yuan2024mars}, and on OLion, which applies the sign operation of Lion \cite{chen2023lion} to the orthogonalized momentum \cite{wang2026olion}.

\section{Methods}\label{sec:methods}

All methods share the same federated protocol. Each of the $N$ clients holds a disjoint set of $n_i$ cars. In round $r$, the server sends the global model $x^r$ to a set $S_r$ of clients, which contains either all clients or, under partial participation, $K$ clients sampled uniformly without replacement. Each client trains the model for $E$ local epochs, that is, $\tau_i \approx E n_i/B$ steps with batch size $B$, and returns the model change $\Delta_i = x_i - x^r$. The server then sets $x^{r+1} = x^r + |S_r|^{-1}\sum_{i \in S_r}\Delta_i$. Gradients are clipped to unit norm, and the learning rate changes between rounds but is constant within a round. The methods differ in the local update and in the state that clients keep across rounds. Algorithm~\ref{alg:spectral} summarizes the spectral methods.

The spectral methods treat weight matrices separately from the remaining parameters. For a weight matrix $W \in \mathbb{R}^{m \times n}$, client $i$ maintains a momentum $M_i$ and a control variate $C_i$ across rounds, and the server maintains the average $C$ of the control variates. At a local step with stochastic gradient $G$, the client updates
\begin{equation}\label{eq:spectral}
M_i \leftarrow \beta M_i + (1-\beta)\,\tilde G, \qquad
D = M_i - C_i + C, \qquad
W \leftarrow (1 - \eta\lambda)\,W - \eta\,\mathcal{U}(D),
\end{equation}
where $\beta$ is the momentum coefficient, $\eta$ the learning rate and $\lambda$ the weight decay. FedMuon uses $\tilde G = G$ and $\mathcal{U}(D) = 0.2\sqrt{\max(m,n)}\;\mathrm{NS}(D)$, where $\mathrm{NS}$ denotes the orthogonalization of Muon \cite{jordan2024muon}: five Newton--Schulz iterations that map $D = U\Sigma V^\top$ approximately to $UV^\top$. The factor $0.2\sqrt{\max(m,n)}$ matches the root-mean-square entry of the update to that of AdamW, so that a single learning rate applies to matrices of all shapes \cite{liu2025muonscalable}. FedMARS-M keeps this update and feeds the momentum with $\tilde G = G + \gamma\frac{\beta}{1-\beta}\,(G - G_i')$, clipped to unit norm, where $G_i'$ is the gradient that the client computed at its previous step (at the first step of a round, its last gradient of the previous round). The correction term reduces the variance of the momentum \cite{yuan2024mars}. Exact MARS evaluates $G_i'$ at the previous weights on the current batch, which requires a second forward and backward pass at every step. We use the approximate form proposed by \citet{yuan2024mars}, which reuses the stored gradient and therefore adds no computation. Its authors report that the exact form is only marginally better. FedOLion uses $\tilde G = G$ and the OLion update $\mathcal{U}(D) = 0.2\,\mathrm{sign}(\mathrm{NS}(D))$ \cite{wang2026olion}.

The control variates correct client drift. At the end of a round, each participating client sets $C_i$ to the average of its momentum over the round and uploads it together with $\Delta_i$, and the server recomputes $C$ from the latest $C_i$ of all clients. The direction $D$ in \eqref{eq:spectral} thus replaces the average direction of the client with the average direction of all clients, as SCAFFOLD does for gradients \cite{karimireddy2020scaffold}. We apply this correction before the orthogonalization, because orthogonalization is nonlinear and the average of orthogonalized client updates differs from the orthogonalized average \cite{takezawa2025fedmuon}. The remaining parameters, about 1\% of the model, take an Adam step on the same corrected momentum (Appendix~\ref{app:hyper}). With a single client, the method reduces to Muon, which serves as our centralized Muon reference.

% A float, and not breakable, so that the box is never cut by a page break.
\begin{figure}[t]
\begin{algorithm}[breakable=false]{Spectral federated training for one weight matrix $W$}
\label{alg:spectral}
\algrenewcommand{\algorithmicrequire}{\textbf{Input:}}
\algrenewcommand{\algorithmicensure}{\textbf{Initialize:}}
\begin{algorithmic}[1]
\Require rounds $R$, learning rates $\eta_0, \dots, \eta_{R-1}$, momentum $\beta$, weight decay $\lambda$, MARS weight $\gamma$
\Ensure $W^0$, \ $C = 0$, \ $M_i = C_i = G_i' = 0$ for all clients $i$
\For{$r = 0, 1, \dots, R-1$}
  \State server samples the clients $S_r$ and sends them $W^r$ and $C$
  \For{each client $i \in S_r$ in parallel}
    \State $W \gets W^r$, \ $\bar M \gets 0$
    \For{$t = 1, \dots, \tau_i$}
      \State $G \gets$ stochastic gradient at $W$
      \State $\tilde G \gets G + \gamma\frac{\beta}{1-\beta}(G - G_i')$, \ $G_i' \gets G$
      \State $M_i \gets \beta M_i + (1-\beta)\,\tilde G$
      \State $D \gets M_i - C_i + C$
      \State $W \gets (1-\eta_r\lambda)\,W - \eta_r\,\mathcal{U}(D)$
      \State $\bar M \gets \bar M + M_i/\tau_i$
    \EndFor
    \State $C_i \gets \bar M$
    \State send $\Delta_i = W - W^r$ and $C_i$ to the server
  \EndFor
  \State $W^{r+1} \gets W^r + |S_r|^{-1}\sum_{i \in S_r}\Delta_i$
  \State $C \gets N^{-1}\sum_{j=1}^{N} C_j$
\EndFor
\State \Return $W^R$
\end{algorithmic}
\smallskip{\footnotesize FedMuon and FedOLion use $\gamma = 0$. The update is $\mathcal{U}(D) = 0.2\sqrt{\max(m,n)}\;\mathrm{NS}(D)$ for FedMuon and FedMARS-M and $\mathcal{U}(D) = 0.2\,\mathrm{sign}(\mathrm{NS}(D))$ for FedOLion. Gradients and $\tilde G$ are clipped to unit norm. The average $C$ uses the latest $C_j$ of every client, including clients that did not take part in round $r$.}
\end{algorithm}
\end{figure}

The non-spectral methods update each parameter independently. SCAFFOLD-Adam keeps the momentum and the control variates of FedMuon but replaces the orthogonalized step with an Adam step for all parameters, uses no weight decay, and decays all of its learning rates. FedAvg clients run AdamW \cite{loshchilov2019adamw} and reset the optimizer state at the start of every round. FedLion clients run Lion \cite{chen2023lion}, and the server averages the client momenta together with the models. None of the methods compresses its uploads, so our FedLion transmits full-precision updates rather than the signed updates of the original method \cite{tang2024fedlion}. The centralized AdamW reference is FedAvg with a single client, so it also resets its optimizer state at every epoch.

\section{Experiments}\label{sec:experiments}

\subsection{Setup}\label{sec:setup}

We use the 5\,703 training and 1\,154 test cars of DrivAerNet++ \cite{elrefaie2024drivaernetpp}. Each car is represented by a point cloud of about 414 thousand surface points, and the surrogate predicts the pressure and the wall shear stress at every point. Integrating these fields over the surface gives the drag and lift coefficients \cd\ and \cl\ (Figure~\ref{fig:overview}b). Six clients hold disjoint subsets of the training cars. In the single-class partition, each client holds cars of a single body type: one client holds estatebacks, one notchbacks and four fastbacks, in proportion to their share of the data. This is the most heterogeneous partition, and we use it for tuning. Three further partitions are drawn from a Dirichlet distribution with $\alpha \in \{0.5, 10, 1000\}$ \cite{hsu2019dirichlet}, ranging from strongly heterogeneous to nearly identical clients.

The surrogate is a Transolver \cite{wu2024transolver} with 1.87 million parameters (Appendix~\ref{app:setup}). Each run lasts 100 rounds with a batch size of 32 cars. All methods decay their learning rates to zero with a cosine schedule \cite{loshchilov2017sgdr}, a common choice for transformers \cite{hoffmann2022chinchilla} and for Muon at scale \cite{liu2025muonscalable}. The only exception is the Adam learning rate of the non-matrix parameters in the spectral methods, which stays constant. We do not use learning-rate warm-up \cite{goyal2017accurate}, whose length would otherwise have to be tuned for each method \cite{riabinin2026warmup}. Federated optimizers are often compared at constant learning rates, as in \citet{reddi2021fedopt}, but in our setting the cosine schedule lowered the drag error of every non-spectral method at its selected number of local epochs (Appendix~\ref{app:schedule}). As references, we train two centralized models on all cars, one with Muon and one with AdamW. All runs use seed 42. For the spectral methods and centralized Muon, we repeat the runs with seeds 43 and 44, both at the tuned number of local epochs and at one local epoch, and report the mean over the three seeds in Tables~\ref{tab:headline} and~\ref{tab:heterogeneity} and Figure~\ref{fig:heterogeneity}, followed in Table~\ref{tab:headline} by the standard deviation after $\pm$.

We report the mean absolute error of \cd\ and \cl\ on the test cars in units of $10^{-3}$, averaged over rounds 96 to 100. Averaging over several rounds is necessary because the force errors fluctuate much more than the training loss. Owing to the cancellation in the force integral, small changes in the predicted fields shift the drag considerably, and the errors stabilize only once the learning rate has decayed. In the result tables, cells are shaded from blue (smallest value) to red (largest value) on a logarithmic scale, separately for each metric.

All methods are tuned with the same three-stage protocol (Appendices~\ref{app:hyper} and~\ref{app:ladders}). Stage 1 selects the momentum coefficients with a Bayesian search at reduced resolution. Stage 2 selects the learning rate for each partition at full resolution on a grid of values $1\!\times\!10^{-j}$ and $3\!\times\!10^{-j}$, moving along the grid until a rate outperforms both of its neighbors. Stage 3 evaluates every number of local epochs $E \in \{1, 2, 4, 8, 16\}$ on the single-class partition at the stage-2 learning rate, keeps the best, and re-tunes the learning rate at that number. Stages 2 and 3 select on the mean drag error at rounds 20, 25 and 30, and the lift error is never used for selection. Tuning and evaluation use the same test cars, so the reported errors are not held-out estimates, although the protocol is identical for all methods. Stages 2 and 3 of the non-spectral methods ran on RTX A6000 GPUs and their 100-round runs on A100 GPUs (Appendix~\ref{app:a6000}). The experiments required about 3\,950 GPU hours in total (Appendix~\ref{app:compute}).

\subsection{Accuracy}\label{sec:results}

Table~\ref{tab:headline} compares the methods on the single-class partition, each at its tuned number of local epochs. FedMARS-M matches centralized Muon within the variation across seeds, and FedMuon is 2\% behind it. The non-spectral methods require sixteen local epochs, eight times the local computation of FedMARS-M, and still have higher drag errors: by 17\% for SCAFFOLD-Adam, 32\% for FedLion and 37\% for FedAvg. With centralized data the choice of optimizer matters less, and AdamW is 12\% behind Muon. Part of the gap between the groups is therefore present without federation. Measured against the centralized model with the matching optimizer, federation increases the drag error by 2\% for FedMuon (against Muon), and by 5\% for SCAFFOLD-Adam and 23\% for FedAvg (against AdamW). FedOLion is the weakest spectral method. Its eight local epochs, selected on 30-round runs, bring no gain over 100 rounds, where it reaches $7.90 \pm 0.51$, compared with 7.47 at one local epoch.

\begin{table}[h]{0.7\columnwidth}
\centering
\caption{Single-class partition after 100 rounds: errors of the drag and lift coefficients in units of $10^{-3}$, each method at its tuned number of local epochs $E$.}\label{tab:headline}
\begin{tabular}{@{}llccc@{}}
\toprule
Group & Method & $E$ & \cd\ error & \cl\ error \\
\midrule
Spectral     & FedMARS-M     & 2  & \heatL{40}$\mathbf{6.77} \pm 0.04$ & \heatL{36}$\mathbf{26.3} \pm 0.3$ \\
             & FedMuon       & 4  & \heatL{36}$6.86 \pm 0.08$ & \heatL{36}$26.4 \pm 0.1$ \\
             & FedOLion      & 8  & \heatH{0}$7.90 \pm 0.51$ & \heatL{24}$27.7 \pm 0.2$ \\
\midrule
Non-spectral & SCAFFOLD-Adam & 16 & \heatH{0}$7.94$ & \heatL{32}$27.0$ \\
             & FedLion       & 16 & \heatH{32}$8.94$ & \heatH{16}$33.3$ \\
             & FedAvg        & 16 & \heatH{40}$9.26$ & \heatH{40}$37.1$ \\
\midrule
Centralized  & Muon          & -- & \heatL{40}$6.73 \pm 0.08$ & \heatL{40}$26.0 \pm 0.3$ \\
             & AdamW         & -- & \heatL{12}$7.54$ & \heatL{24}$27.9$ \\
\bottomrule
\end{tabular}
\end{table}

The gap between the two groups is much larger when every method uses one local epoch per round (Figure~\ref{fig:heterogeneity} and Table~\ref{tab:heterogeneity}). The spectral methods reach drag errors between 6.8 and 7.5 on all four partitions, within 11\% of centralized Muon even when each client holds a single body type. The non-spectral methods reach 11 to 18, that is, 1.5 to 2.6 times the error of the spectral methods on the same partition. The only exception is FedAvg at $\alpha = 10$ (32.6), whose learning rate of $3\!\times\!10^{-3}$ won the 30-round search by a small margin. A repeated run on A6000 GPUs reaches 20.7, so the exception is partly due to run-to-run variation.

\begin{figure}[h]
\centering
\includegraphics[width=\textwidth]{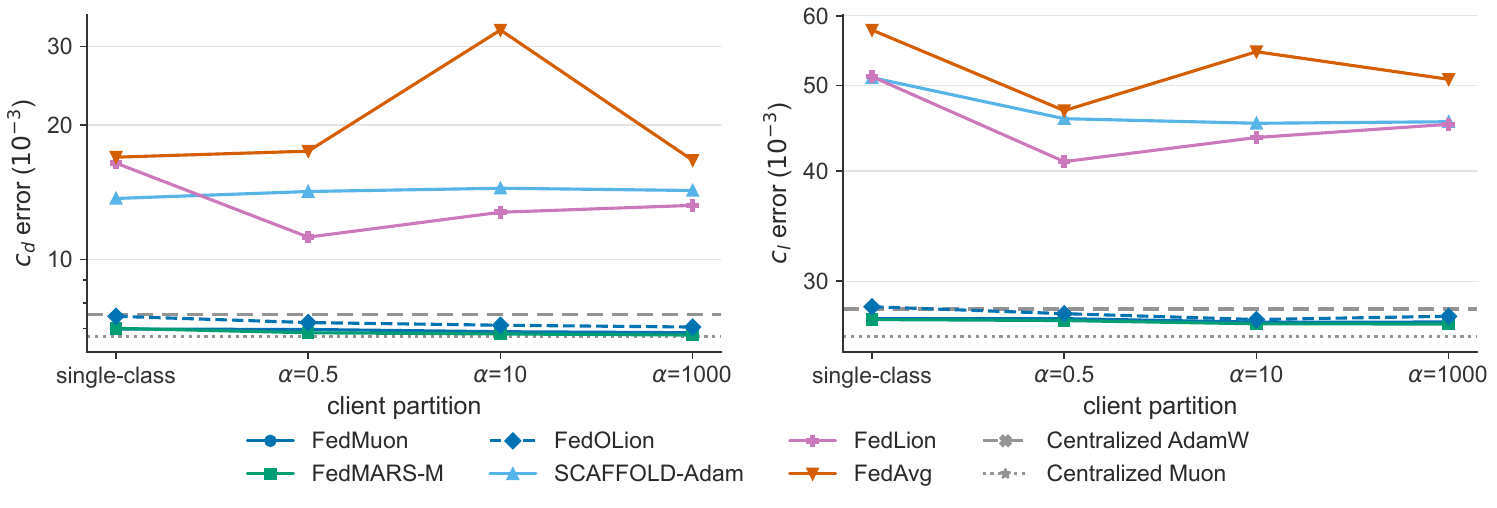}
\caption{Errors on the four client partitions with one local epoch per round, after 100 rounds. Horizontal lines show the centralized models.}\label{fig:heterogeneity}
\end{figure}

\begin{table}[h]
\centering
\caption{One local epoch per round on the four partitions after 100 rounds, errors in units of $10^{-3}$. Values above 17.5 have the same color as 17.5.}\label{tab:heterogeneity}
\small
\setlength{\tabcolsep}{4.5pt}
\begin{tabular}{@{}lrrrrrrrr@{}}
\toprule
 & \multicolumn{4}{c}{\cd\ error} & \multicolumn{4}{c}{\cl\ error} \\
\cmidrule(lr){2-5}\cmidrule(lr){6-9}
Method & single-class & $\alpha = 0.5$ & $\alpha = 10$ & $\alpha = 1000$ & single-class & $\alpha = 0.5$ & $\alpha = 10$ & $\alpha = 1000$ \\
\midrule
FedMARS-M     & \heatL{36}7.02 & \heatL{40}6.86 & \heatL{40}6.82 & \heatL{40}6.78 & \heatL{40}27.1 & \heatL{40}27.1 & \heatL{40}26.8 & \heatL{40}26.8 \\
FedMuon       & \heatL{36}6.99 & \heatL{36}6.98 & \heatL{40}6.90 & \heatL{40}6.86 & \heatL{40}27.2 & \heatL{40}27.2 & \heatL{40}27.0 & \heatL{40}27.0 \\
FedOLion      & \heatL{32}7.47 & \heatL{36}7.23 & \heatL{36}7.13 & \heatL{36}7.07 & \heatL{36}28.1 & \heatL{36}27.6 & \heatL{40}27.1 & \heatL{36}27.4 \\
\midrule
SCAFFOLD-Adam & \heatH{20}13.7 & \heatH{24}14.2 & \heatH{24}14.4 & \heatH{24}14.3 & \heatH{28}51.1 & \heatH{16}45.9 & \heatH{16}45.3 & \heatH{16}45.5 \\
FedLion       & \heatH{36}16.4 & \heatH{4}11.2 & \heatH{12}12.8 & \heatH{16}13.2 & \heatH{28}51.2 & \heatH{4}41.0 & \heatH{12}43.7 & \heatH{16}45.2 \\
FedAvg        & \heatH{36}17.0 & \heatH{40}17.5 & \heatH{40}32.6 & \heatH{36}16.7 & \heatH{40}57.8 & \heatH{20}46.8 & \heatH{36}54.6 & \heatH{28}50.8 \\
\bottomrule
\end{tabular}
\end{table}

The training loss decreases for all methods, but the force errors do not follow it (Figure~\ref{fig:convergence}). The drag error of FedMuon rises from 13 at round 50 to 34 at round 70 and returns to 7 at round 100 while its training loss continues to decrease, and FedOLion shows the same behavior. We attribute these fluctuations to the cancellation in the force integral. They vanish as the learning rate decays.

\begin{figure}[h]
\centering
\includegraphics[width=\textwidth]{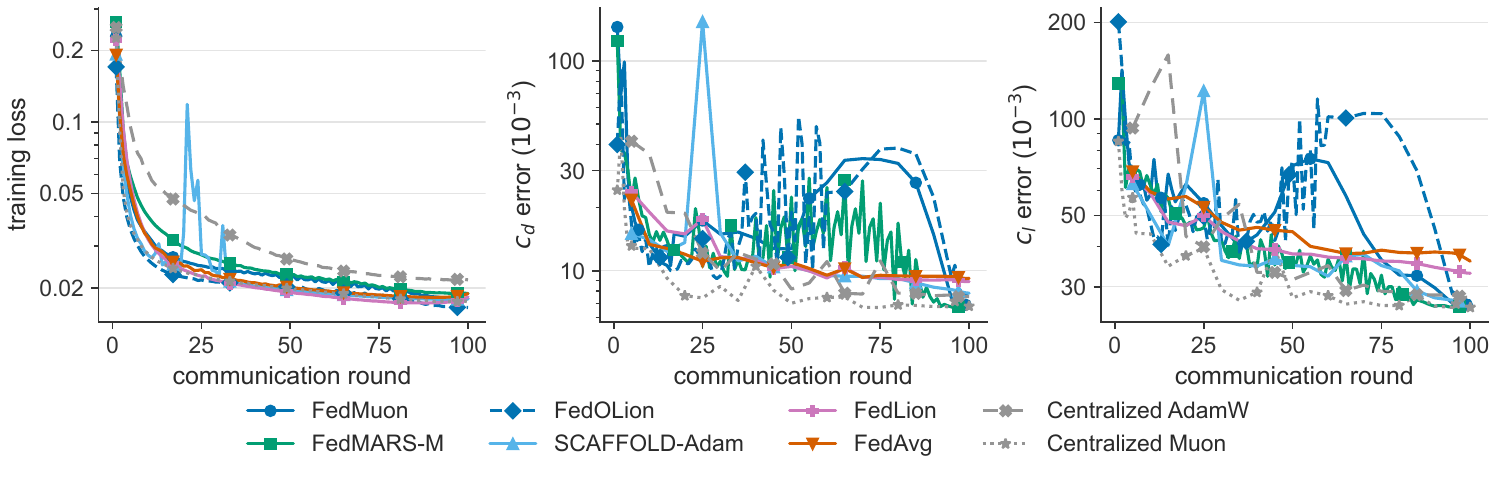}
\caption{Training loss and force errors of the global model over the rounds on the single-class partition, for the runs of Table~\ref{tab:headline}.}\label{fig:convergence}
\end{figure}

FedMARS-M differs from FedMuon in the MARS term and in its separately tuned Adam learning rate. To isolate the effect of the term, we train FedMARS-M at one local epoch on the single-class partition with $\gamma = 0$, $0.01$ and $0.04$ and all other settings fixed, which yields drag errors of 7.43, 6.93 and 6.94. The 7\% improvement is only about twice the standard deviation of FedMARS-M across seeds at one local epoch (0.26). Averaged over three seeds, FedMARS-M and FedMuon perform equally at one local epoch (7.02 and 6.99), and FedMARS-M is slightly better at its tuned number of local epochs (6.77 compared with 6.86). The MARS term therefore provides at most a small gain.

\subsection{Local epochs}\label{sec:epochs}

Figure~\ref{fig:epochs} and Table~\ref{tab:epochs} report the drag error after 30 rounds as a function of the number of local epochs per round. The two groups prefer very different numbers of local epochs.

\noindent
\begin{minipage}[t]{0.44\textwidth}
\vspace{0pt}
FedMuon and FedMARS-M perform best with two to four local epochs and degrade quickly beyond that. At eight epochs, the error of FedMuon is 45\% higher than at four, and that of FedMARS-M is 75\% higher than at two. FedOLion performs best at eight epochs, where it attains the lowest 30-round error of all methods. The non-spectral methods improve up to sixteen epochs. With their learning rates re-tuned at sixteen epochs in stage 3, they reach 9.2 to 9.6, on par with FedMARS-M at two epochs but with eight times the local computation.
\end{minipage}\hfill
\begin{minipage}[t]{0.53\textwidth}
\vspace{0pt}
\begin{braintablebox}[before skip=0pt, after skip=0pt]{\linewidth}
\centering
\captionof{table}{Drag error after 30 rounds for each number of local epochs $E$ on the single-class partition, in units of $10^{-3}$. Bold marks the selected $E$.}\label{tab:epochs}
\small
\setlength{\tabcolsep}{4pt}
\begin{tabular}{@{}lrrrrr@{}}
\toprule
Method & $E = 1$ & 2 & 4 & 8 & 16 \\
\midrule
FedMARS-M     & \heatL{24}10.8 & \heatL{36}\textbf{9.26} & \heatL{36}9.29 & \heatH{4}16.2 & \heatL{16}12.0 \\
FedMuon       & \heatL{20}11.6 & \heatL{28}10.3 & \heatL{32}\textbf{9.63} & \heatL{4}14.0 & \heatH{20}20.2 \\
FedOLion      & \heatL{12}12.7 & \heatL{36}9.13 & \heatL{40}8.90 & \heatL{40}\textbf{8.66} & \heatL{8}13.2 \\
\midrule
SCAFFOLD-Adam & \heatH{12}17.4 & \heatH{16}19.0 & \heatL{4}13.9 & \heatL{20}11.7 & \heatL{24}\textbf{10.7} \\
FedLion       & \heatH{40}25.9 & \heatH{28}22.4 & \heatH{8}16.3 & \heatL{8}13.7 & \heatL{28}\textbf{10.3} \\
FedAvg        & \heatH{16}18.7 & \heatH{16}19.1 & \heatH{16}18.3 & \heatL{4}14.4 & \heatL{20}\textbf{11.6} \\
\bottomrule
\end{tabular}
\end{braintablebox}
\end{minipage}

On the Dirichlet partitions, where the clients are more similar, the spectral methods tolerate more local computation, and the best number of local epochs for FedMuon and FedMARS-M lies between four and sixteen. FedOLion performs best at eight epochs on all three partitions and is the only spectral method that still trains at sixteen epochs for $\alpha = 0.5$.

\begin{figure}[h]
\centering
\includegraphics[width=\textwidth]{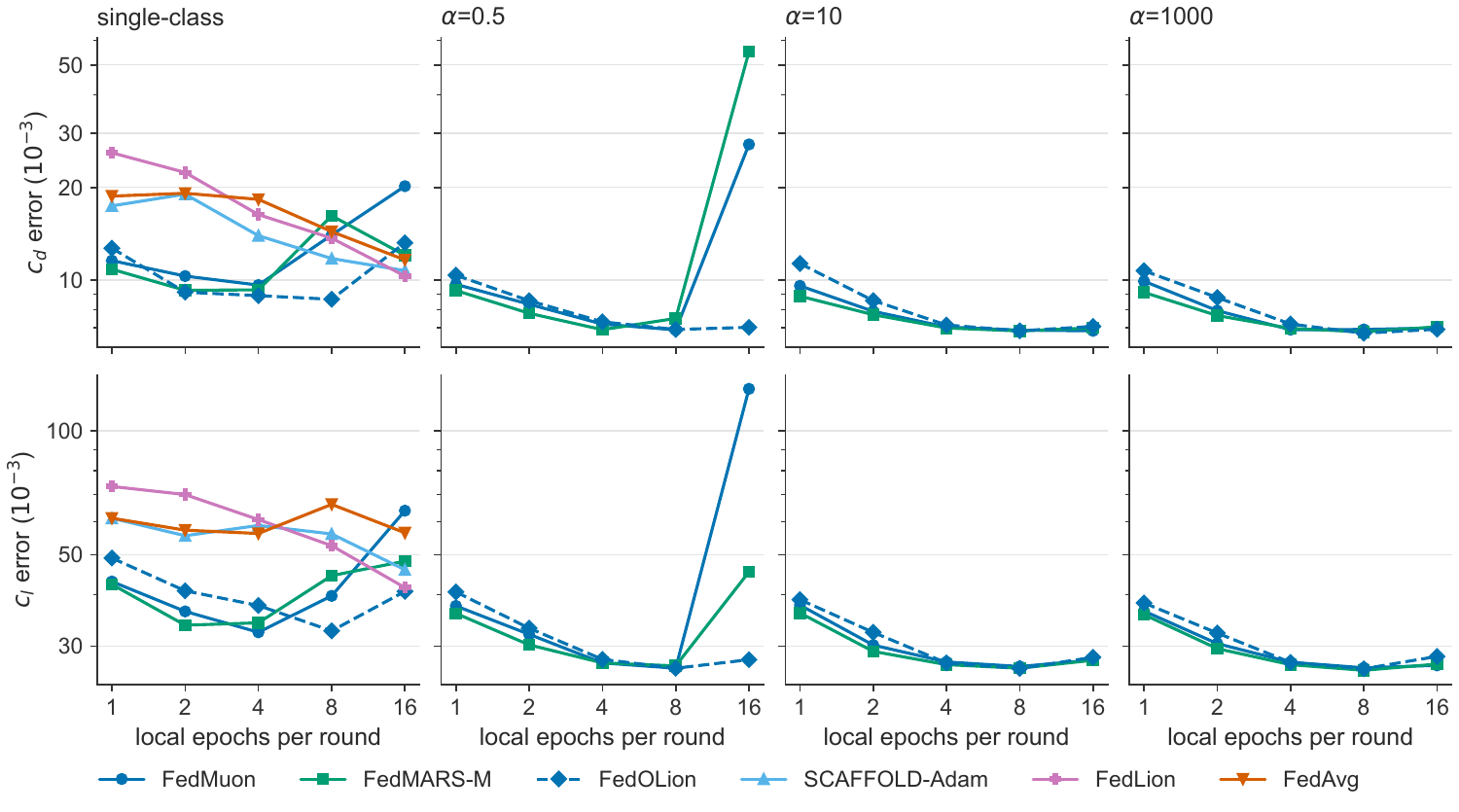}
\caption{Errors after 30 rounds as a function of the number of local epochs per round, with drag in the top row and lift in the bottom row. On the Dirichlet partitions, only the spectral methods were run.}\label{fig:epochs}
\end{figure}

\subsection{Partial participation}\label{sec:participation}

Table~\ref{tab:participation} reports results on the single-class partition when only $K$ of the six clients participate in each round. FedMARS-M and FedMuon lose no accuracy with three clients per round and lose 5 to 7\% in drag error with one client per round. For every $K$, the best non-spectral method has a higher drag error than FedMARS-M. SCAFFOLD-Adam diverges with one and with two clients per round. In these cases, its updates are no longer averaged over six clients and its control variates can be many rounds old, and the training loss becomes NaN after about 36 rounds. This agrees with earlier reports that SCAFFOLD performs poorly under client sampling \cite{li2022silos}. Figures~\ref{fig:participation} and~\ref{fig:participation_comm} show the corresponding training curves. Measured in client uploads, one client per round reaches a given training loss at the lowest cost, whereas measured in rounds, full participation converges fastest.

\begin{table}[h]{0.8\columnwidth}
\centering
\caption{Single-class partition with $K$ of the six clients per round, after 100 rounds, errors in units of $10^{-3}$. NaN marks a diverged run.}\label{tab:participation}
\small
\setlength{\tabcolsep}{5pt}
\begin{tabular}{@{}lcrrrrrrrr@{}}
\toprule
 &  & \multicolumn{4}{c}{\cd\ error} & \multicolumn{4}{c}{\cl\ error} \\
\cmidrule(lr){3-6}\cmidrule(lr){7-10}
Method & $E$ & $K = 1$ & 2 & 3 & 6 & $K = 1$ & 2 & 3 & 6 \\
\midrule
FedMARS-M     & 2 & \heatL{28}7.21 & \heatL{40}6.65 & \heatL{40}6.69 & \heatL{40}6.73 & \heatL{32}27.1 & \heatL{40}26.2 & \heatL{36}26.3 & \heatL{40}26.0 \\
FedMuon       & 4 & \heatL{28}7.30 & \heatL{32}7.07 & \heatL{36}6.85 & \heatL{36}6.93 & \heatH{0}34.3 & \heatL{40}26.1 & \heatL{40}25.9 & \heatL{36}26.4 \\
FedOLion      & 8 & \heatL{8}8.59 & \heatL{24}7.47 & \heatL{32}7.02 & \heatL{12}8.45 & \heatH{36}43.0 & \heatL{32}27.5 & \heatL{28}27.8 & \heatL{28}27.7 \\
\midrule
SCAFFOLD-Adam & 16 & \heatNA NaN & \heatNA NaN & \heatL{16}8.01 & \heatL{16}7.94 & \heatNA NaN & \heatNA NaN & \heatL{24}28.7 & \heatL{32}27.0 \\
FedLion       & 16 & \heatH{40}12.6 & \heatL{12}8.32 & \heatH{0}9.27 & \heatL{4}8.94 & \heatH{40}44.3 & \heatL{12}30.9 & \heatH{4}34.6 & \heatL{4}33.3 \\
FedAvg        & 16 & \heatL{24}7.59 & \heatH{16}10.3 & \heatH{8}9.80 & \heatH{0}9.26 & \heatH{8}36.1 & \heatL{4}33.3 & \heatH{4}34.8 & \heatH{12}37.1 \\
\bottomrule
\end{tabular}
\end{table}

\begin{figure}[tbp]
\centering
\includegraphics[width=\textwidth]{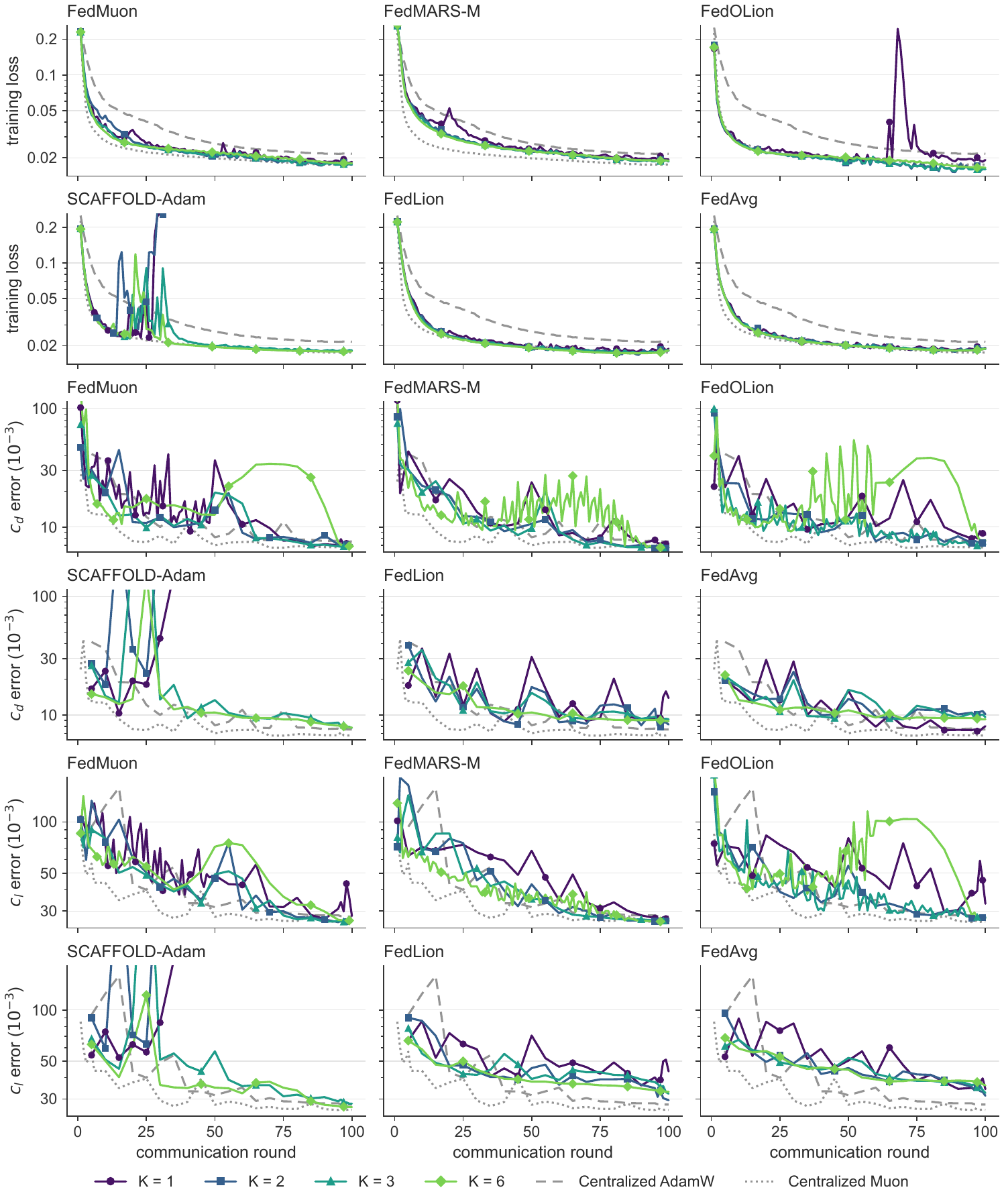}
\caption{Partial participation on the single-class partition: training loss and force errors over the rounds for $K$ clients per round.}\label{fig:participation}
\end{figure}

\begin{figure}[tbp]
\centering
\includegraphics[width=\textwidth]{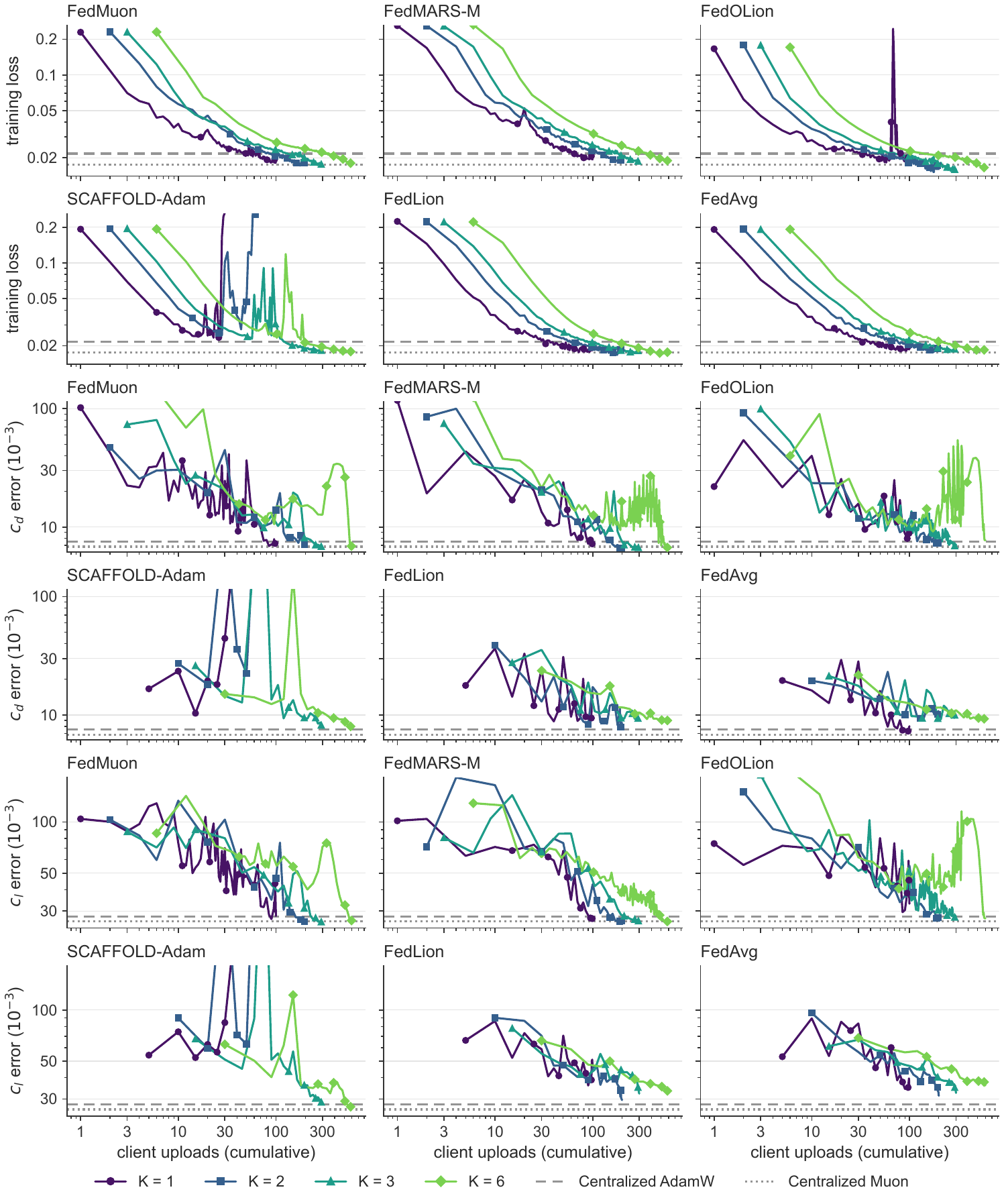}
\caption{The runs of Figure~\ref{fig:participation} as a function of the number of client uploads.}\label{fig:participation_comm}
\end{figure}

\section{Why spectral methods need fewer local steps}\label{sec:why}

In Section~\ref{sec:epochs}, the spectral methods performed best with two to eight local epochs per round, whereas the non-spectral methods still improved at sixteen. We now give a heuristic explanation of both observations based on the singular values of the weight gradients.

Consider a weight matrix $W \in \mathbb{R}^{m \times n}$ with gradient $G = U\Sigma V^\top$, and let $\Delta W$ be a step, so that the loss changes by $\langle G, \Delta W\rangle$ to first order. Since $\|\Delta W x\| \le \|\Delta W\|_2\,\|x\|$ for every input $x$ of the layer, the spectral norm of $\Delta W$ bounds the change of the layer output, and we compare steps of equal spectral norm $\eta$ (for FedMuon and FedMARS-M, $\eta$ is the learning rate multiplied by $0.2\sqrt{\max(m,n)}$). Among such steps, the orthogonalized step $\Delta W = -\eta\,UV^\top$ achieves the largest decrease of the linearized loss, $\eta\|G\|_*$ \cite{bernstein2024anthology,kovalev2025trust}, whereas the gradient step $\Delta W = -\eta\,G/\sigma_{\max}(G)$ achieves $\eta\|G\|_F^2/\sigma_{\max}(G)$. We define the gain of orthogonalization as the ratio of the two,
\[
g(G) = \frac{\|G\|_*\,\sigma_{\max}(G)}{\|G\|_F^2} \;\ge\; 1 .
\]
The gain equals one if and only if all non-zero singular values of $G$ are equal, and it is large when one singular value dominates many smaller but non-negligible ones. For example, if a gradient of rank $k$ has $k-1$ equal smaller singular values, then $g$ is largest at a stable rank $\mathrm{sr}(G) = \|G\|_F^2/\sigma_{\max}^2(G)$ slightly below two, where it equals $(\sqrt{k}+1)/2$.

Figure~\ref{fig:spectrum} reports these quantities for the 42 weight matrices of the surrogate on the single-class partition, computed from 16 cars per client after 100 rounds of training. For models trained with any of the non-spectral methods or with centralized AdamW, the median stable rank is between 2.0 and 2.5, although the matrices have rank 32 or 256, and the median gain is between 2.2 and 2.3. At equal spectral norm, one orthogonalized step therefore achieves roughly the first-order decrease of two gradient steps.

\begin{figure}[h]
\centering
\includegraphics[width=\textwidth]{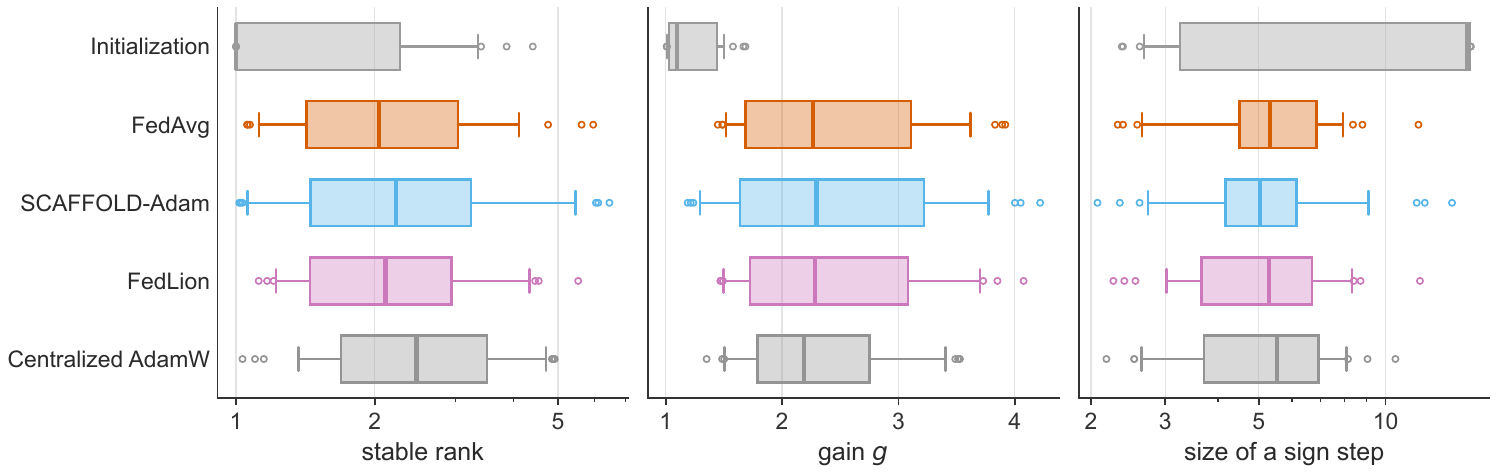}
\caption{Gradient spectra of the 42 weight matrices on the single-class partition: stable rank, gain $g$ of an orthogonalized step, and relative spectral norm $\rho$ of a sign step.}\label{fig:spectrum}
\end{figure}

The non-spectral methods, however, do not take gradient steps. Lion steps along the sign of a momentum, and Adam steps are close to sign steps early in training. The learning rate of these methods therefore controls the magnitude of the entries of the step, that is, its Frobenius norm, rather than its spectral norm. A sign matrix has Frobenius norm $\sqrt{mn}$, while an orthogonal matrix rescaled to the same Frobenius norm has spectral norm $\sqrt{\max(m,n)}$. At equal Frobenius norm, a sign step therefore has a larger spectral norm than an orthogonalized step, by the factor
\[
\rho(G) = \frac{\|\mathrm{sign}(G)\|_2}{\sqrt{\max(m,n)}} \;\in\; \bigl[1, \sqrt{\min(m,n)}\bigr],
\]
whose upper bound is attained when the signs of $G$ form a rank-one pattern, as is typical for gradients of low stable rank. On the trained models, the median of $\rho$ is between 5 and 5.5 (Figure~\ref{fig:spectrum}, right).

To keep the change of the layer outputs within the same bound, an element-wise method must therefore take steps several times smaller in Frobenius norm than a spectral method, and it needs correspondingly more steps. The tuned runs are consistent with this. Over one local epoch, a FedMuon client moves by 12.2 in Frobenius norm, a FedLion client by 4.3, and FedAvg and SCAFFOLD-Adam clients by 0.6 and 0.5 (Table~\ref{tab:drift}). FedOLion takes the sign of an orthogonalized direction, so the argument applies to it only in part, which is consistent with its intermediate number of eight local epochs.

It remains to explain why additional local epochs hurt the spectral methods. The norm of an orthogonalized step does not depend on the norm of the gradient, so a client does not slow down as it approaches a stationary point of its local objective. After $\tau_i$ local steps it can be as far as $\eta\tau_i$ from the global model in spectral norm, and Table~\ref{tab:drift} shows that the spectral clients nearly attain this linear growth: each doubling of $E$ up to $E = 8$ increases their movement per round by a factor of 1.6 to 2.1, so their successive steps point in similar directions. At the same time, the fraction of the movement retained after averaging falls from 60--65\% at $E = 1$ to 35--42\% at $E = 8$, so most of the additional movement is client-specific. A fixed step size alone does not produce this behavior. The sign steps of Lion also have a fixed size, but from $E = 1$ to $E = 16$ the movement of FedLion grows only by a factor of 3.8, and that of SCAFFOLD-Adam and FedAvg by 3.9 and 2.5. For the non-spectral methods, 76 to 91\% of the movement is retained for every $E$.

\begin{table}[h]{0.8\columnwidth}
\centering
\caption{Client drift at round 5 of the runs of Table~\ref{tab:epochs}: the mean Frobenius norm of the client updates over one round (all parameters), and the norm of their average as a fraction of it.}\label{tab:drift}
\small
\setlength{\tabcolsep}{5pt}
\begin{tabular}{@{}lrrrrrrrrrr@{}}
\toprule
 & \multicolumn{5}{c}{Movement per round} & \multicolumn{5}{c}{Kept by averaging (\%)} \\
\cmidrule(lr){2-6}\cmidrule(lr){7-11}
Method & $E = 1$ & 2 & 4 & 8 & 16 & $E = 1$ & 2 & 4 & 8 & 16 \\
\midrule
FedMARS-M     & \heatH{8}12.3 & \heatH{16}23.4 & \heatH{28}46.0 & \heatH{32}74.3 & \heatH{40}102 & \heatH{4}60 & \heatL{12}49 & \heatL{24}42 & \heatL{36}37 & \heatL{32}38 \\
FedMuon       & \heatH{8}12.2 & \heatH{16}25.1 & \heatH{28}46.1 & \heatH{36}84.0 & \heatH{40}116 & \heatH{4}60 & \heatL{12}49 & \heatL{24}43 & \heatL{40}35 & \heatL{36}36 \\
FedOLion      & \heatL{4}5.85 & \heatH{4}10.5 & \heatH{12}17.7 & \heatH{20}30.7 & \heatH{28}52.2 & \heatH{12}65 & \heatL{4}55 & \heatL{12}48 & \heatL{24}42 & \heatL{36}37 \\
\midrule
SCAFFOLD-Adam & \heatL{40}0.53 & \heatL{32}0.88 & \heatL{28}1.28 & \heatL{24}1.45 & \heatL{20}2.08 & \heatH{32}83 & \heatH{40}91 & \heatH{40}90 & \heatH{36}87 & \heatH{40}90 \\
FedLion       & \heatL{8}4.30 & \heatL{4}6.60 & \heatH{4}9.72 & \heatH{8}12.4 & \heatH{12}16.4 & \heatH{40}91 & \heatH{40}89 & \heatH{36}85 & \heatH{28}79 & \heatH{24}76 \\
FedAvg        & \heatL{40}0.60 & \heatL{32}0.80 & \heatL{32}1.00 & \heatL{28}1.17 & \heatL{24}1.52 & \heatH{36}87 & \heatH{40}91 & \heatH{40}89 & \heatH{32}83 & \heatH{28}80 \\
\bottomrule
\end{tabular}
\end{table}

These measurements are consistent with the explanation but do not establish it. In particular, the spectra were computed on models trained with the non-spectral methods and with centralized AdamW, and we did not vary the stable rank of the gradients independently of the optimizer.

\section{Discussion}\label{sec:discussion}

Our results indicate that the choice of optimizer largely determines how close a federated aerodynamic surrogate comes to one trained on pooled data. With FedMuon or FedMARS-M, the shared surrogate predicts drag and lift almost as accurately as centralized Muon, even when each client holds a single body type, and it requires only two to four local epochs per round. Compared with FedAvg, the spectral methods double the upload in every round, as SCAFFOLD-Adam and FedLion do.

\section{Limitations}\label{sec:limitations}

Our study is limited to one dataset, one architecture and six clients. The same test cars are used for tuning and evaluation, and the non-spectral methods were run with a single seed. We considered at most sixteen local epochs, although the non-spectral methods were still improving at that point. The number of local epochs is selected on 30-round runs and need not be optimal over 100 rounds, as the case of FedOLion shows. Our clients are partitions of a single public dataset, whereas real partners would also differ in meshes, solvers and flow conditions. The strongest non-spectral configuration, SCAFFOLD-Adam with sixteen local epochs, is fragile: on A6000 GPUs it diverges for every number of participating clients (Appendix~\ref{app:a6000}). Finally, the explanation in Section~\ref{sec:why} is heuristic.

\section{Future work}\label{sec:future}

The most immediate next step is a second architecture. Our implementation runs the same protocol for any model that maps a point cloud to surface pressure and wall shear stress, and we plan to evaluate OmniArch \cite{chen2025omniarch}, which combines a Fourier encoder with a transformer backbone and differs substantially from Transolver. This experiment would indicate whether the low stable rank of the gradients, and with it the advantage of the spectral methods, is a property of the task or of the architecture. We also plan to run the non-spectral methods with additional seeds and to evaluate all final models on the validation split of DrivAerNet++, which we did not use. On the methodological side, compressing the control variates, as has been done for model updates \cite{konecny2016federated}, would reduce the communication cost of the spectral methods, and a convergence analysis of orthogonalized updates with local steps could indicate how many local epochs they can afford. In the longer term, the methods should be evaluated with industrial partners, whose data differ in meshes, solvers and flow conditions, and on aerodynamic data beyond passenger cars.

\section{Conclusion}\label{sec:conclusion}

We studied federated training of a car aerodynamics surrogate across clients that cannot share their designs. On DrivAerNet++ with one body type per client, FedMuon and FedMARS-M reach a drag error within 2\% of centralized Muon with two to four local epochs per round, whereas FedAvg, SCAFFOLD-Adam and FedLion require sixteen local epochs and remain 17 to 37\% worse than FedMARS-M. We relate this gap to the low stable rank of the weight gradients, a regime in which orthogonalized steps make the most progress per step and element-wise steps must remain small. These results indicate that spectral optimizers are a suitable choice for federated training of aerodynamic surrogates.

% the bibliography (printed by \end{mainpart}) starts on a new page
\clearpage
\end{mainpart}

\begin{appendixpart}

\section{Training setup}\label{app:setup}

Table~\ref{tab:setup} lists the settings shared by all methods, and Table~\ref{tab:partitions} describes the four client partitions.

\begin{table}[h]
\centering
\caption{Training setup shared by all methods.}\label{tab:setup}
\small
\begin{tabular*}{\linewidth}{@{\extracolsep{\fill}}lp{0.72\linewidth}@{}}
\toprule
Setting & Value \\
\midrule
Data & DrivAerNet++: 5\,703 training and 1\,154 test cars, about 414 thousand surface points per car (330 to 513 thousand), 125 GB \\
Input and output & point coordinates and surface normals in, pressure and wall shear stress at every point out \\
Model & Transolver: 5 blocks, width 256, 8 heads, 32 slices, MLP ratio 1, dropout 0.2, GELU. 1.87M parameters, of which 1.85M in 42 weight matrices \\
Loss & sum of the mean squared errors of the normalized pressure and wall shear stress \\
Points per car & 50 thousand sampled per training pass, all points at test time \\
Clients & 6, and all of them train in every round unless $K$ is stated \\
Batch & 32 cars, as two accumulated micro-batches of 16 \\
Rounds & 100 (results), 30 (tuning stages 2 and 3), 20 (stage 1) \\
Local epochs $E$ & tuned over $\{1, 2, 4, 8, 16\}$ \\
Server & plain average of the client models (equal weights), server learning rate 1 \\
Schedule & cosine decay to zero over the run: all rates of the non-spectral methods, the matrix rate of the spectral methods (their Adam rate stays constant) \\
Gradient clipping & norm 1 \\
Precision & bf16 mixed precision \\
Evaluation & every fifth round and rounds 96 to 100, on all test cars and points. The side-force coefficient, zero by symmetry, is checked in every run \\
Reported score & mean over rounds 96 to 100 of the \cd\ and \cl\ mean absolute errors \\
Tuning score & stages 2 and 3: mean \cd\ error at rounds 20, 25 and 30. Stage 1: mean of the last five evaluations, with \cl\ and the field errors checked for outliers \\
Seed & 42, and also 43 and 44 for the spectral methods with their tuned number of local epochs and with one local epoch, and for the centralized Muon model \\
\bottomrule
\end{tabular*}
\end{table}

\begin{table}[h]{0.65\columnwidth}
\centering
\caption{Client partitions of the 5\,703 training cars: the smallest, mean and largest number of cars per client, and the mean normalized entropy of the car categories within a client (0.82 for the whole training set).}\label{tab:partitions}
\begin{tabular*}{\linewidth}{@{\extracolsep{\fill}}lccc@{}}
\toprule
Partition & Clients & Cars per client & Entropy \\
\midrule
single-class & 6 & 917 / 950.5 / 1074 & 0.00 \\
Dirichlet $\alpha = 0.5$ & 6 & 554 / 950.5 / 1954 & 0.56 \\
Dirichlet $\alpha = 10$ & 6 & 700 / 950.5 / 1231 & 0.81 \\
Dirichlet $\alpha = 1000$ & 6 & 943 / 950.5 / 959 & 0.82 \\
\bottomrule
\end{tabular*}
\end{table}

\section{Hyperparameters}\label{app:hyper}

Table~\ref{tab:hyper} lists the selected hyperparameters. The momentum coefficients, the Adam learning rate of the spectral methods and the second momentum coefficient of Lion are selected in stage 1. The learning rates are selected in stage 2 (per partition, at $E = 1$) and re-checked in stage 3 at the selected $E$. Table~\ref{tab:spaces} gives the stage-1 search spaces. Stage 1 runs at reduced fidelity (6 thousand points per car, 20 rounds and 150 test cars) and evaluates 24 distinct configurations per method without pruning. FedMARS-M has its own 24 configurations at $\gamma = 0.01$ and a separate search over the five values of $\gamma$ in Table~\ref{tab:spaces}. The search spaces of FedAvg, SCAFFOLD-Adam and centralized AdamW contain only 25 configurations each, and for them the sampler reached its limit of 240 trials after 17 or 18 distinct configurations. For FedLion, the stage-1 winner under the cosine schedule did not transfer to full resolution, so we repeated stage 2 with the momentum coefficients of the constant-rate search and kept the better configuration.

In the spectral methods, the parameters that are not weight matrices (biases, normalization parameters, the input embedding and the output head) are updated with the Adam step $V \leftarrow \beta_2 V + (1-\beta_2)\,D \odot D$, $\theta \leftarrow \theta - \eta_{\mathrm{A}}\, D/(\sqrt{V} + \epsilon)$ applied to the corrected momentum $D$, with $\epsilon = 10^{-8}$, no bias correction and a constant learning rate $\eta_{\mathrm{A}}$. The FedLion step is $\eta\,\mathrm{sign}(\beta_1 M_i + (1-\beta_1)\,G)$ with decoupled weight decay, followed by the momentum update $M_i \leftarrow \beta_2 M_i + (1-\beta_2)\,G$.

\begin{table}[h]
\centering
\caption{Selected hyperparameters, with learning rates written $a$e$b$ for $a \times 10^{b}$. Rates that differ between partitions are listed for single-class, $\alpha = 0.5$, 10 and 1000.}\label{tab:hyper}
\footnotesize
\setlength{\tabcolsep}{3pt}
\begin{tabular*}{\linewidth}{@{\extracolsep{\fill}}lclllll@{}}
\toprule
Method & $E$ & LR at $E = 1$ & LR at tuned $E$ & Adam LR & $\beta_1$ / $\beta_2$ & Weight decay \\
\midrule
FedMuon & 4 & 3e-3 & 3e-3 & 3e-5 & 0.9 / 0.95 & 0.1 (matrices) \\
FedMARS-M & 2 & 3e-3 & 3e-3 & 1e-5 & 0.9 / 0.95 & 0.1 (matrices) \\
FedOLion & 8 & 1e-3 & 1e-3 & 3e-6 & 0.95 / 0.95 & 0.1 (matrices) \\
Centralized Muon & -- & 1e-3 & -- & 1e-5 & 0.95 / 0.95 & 0.1 (matrices) \\
\midrule
FedAvg (AdamW) & 16 & 1e-4, 1e-3, 3e-3, 1e-4 & 3e-4 & -- & 0.7 / 0.999 & 0.01 \\
SCAFFOLD-Adam & 16 & 3e-5 & 1e-4 & -- & 0.8 / 0.95 & -- \\
FedLion & 16 & 3e-4 & 1e-4 & -- & 0.8 / 0.95 & 0.01 \\
Centralized AdamW & -- & 1e-3 & -- & -- & 0.8 / 0.999 & 0.01 \\
\bottomrule
\end{tabular*}
\par\medskip{\footnotesize $\beta_1$ is the momentum $\beta$ of the client update, and $\beta_2$ is the second-moment rate of the Adam steps or, for FedLion, its second momentum coefficient. Spectral methods: 5 Newton--Schulz iterations $X \leftarrow aX + (bA + cA^2)X$ with $A = XX^\top$ and $(a, b, c) = (3.4445, -4.7750, 2.0315)$, started from $D/\|D\|_F$. The orthogonalized step is scaled by $0.2\sqrt{\max(m, n)}$, and the control variates come from the mean client momentum. FedMARS-M: MARS weight $\gamma = 0.01$, correction clipped at norm 1. FedOLion: the sign of the orthogonalized direction. FedLion averages the client weights and momenta.}
\end{table}

\begin{table}[h]
\centering
\caption{Stage-1 search spaces. Size is the number of configurations in the space, and searched the share of them that was evaluated.}\label{tab:spaces}
\footnotesize
\begin{tabular*}{\linewidth}{@{\extracolsep{\fill}}lp{0.5\linewidth}cc@{}}
\toprule
Methods & Searched values & Size & Searched \\
\midrule
spectral & matrix LR \{3e-4, 1e-3, 3e-3, 1e-2, 3e-2\}, Adam LR \{3e-6, 1e-5, 3e-5, 1e-4, 3e-4\}, $\beta_1$ \{0.7, 0.8, 0.9, 0.95, 0.99\} & 125 & 19\% \\
FedAvg, centralized AdamW & LR \{1e-5, 3e-5, 1e-4, 3e-4, 1e-3\}, $\beta_1$ as above & 25 & 68\%, 72\% \\
SCAFFOLD-Adam & LR \{1e-6, 3e-6, 1e-5, 3e-5, 1e-4\}, $\beta_1$ as above & 25 & 72\% \\
FedLion & LR \{1e-5, 3e-5, 1e-4, 3e-4, 1e-3\}, $\beta_1$ as above, $\beta_2$ \{0.9, 0.95, 0.99, 0.995\} & 100 & 24\% \\
FedMARS-M, $\gamma$ & MARS weight $\gamma$ \{0.01, 0.025, 0.04, 0.1, 1\} at FedMARS-M's selection. The value 0.04 scored 0.7\% better than the reference 0.01, a difference within noise, so we kept 0.01 & 5 & 100\% \\
\bottomrule
\end{tabular*}
\end{table}

\section{Learning-rate ladders}\label{app:ladders}

Figure~\ref{fig:ladders} shows all stage-2 searches, that is, the 30-round drag error as a function of the learning rate for each method and partition at one local epoch. Each spectral method selects the same learning rate on every partition ($3\!\times\!10^{-3}$ for FedMuon and FedMARS-M, $10^{-3}$ for FedOLion), as do SCAFFOLD-Adam ($3\!\times\!10^{-5}$) and FedLion ($3\!\times\!10^{-4}$). Only the learning rate of FedAvg depends on the partition. It is $10^{-4}$ on the single-class and $\alpha = 1000$ partitions and $10^{-3}$ or $3\!\times\!10^{-3}$ on the other two, and the rate $10^{-2}$ diverges. Figure~\ref{fig:heterogeneity30} compares the methods at their selected learning rates after 30 rounds. The two groups separate in the same way as after 100 rounds (Figure~\ref{fig:heterogeneity}).

\begin{figure}[h]
\centering
\includegraphics[width=\textwidth]{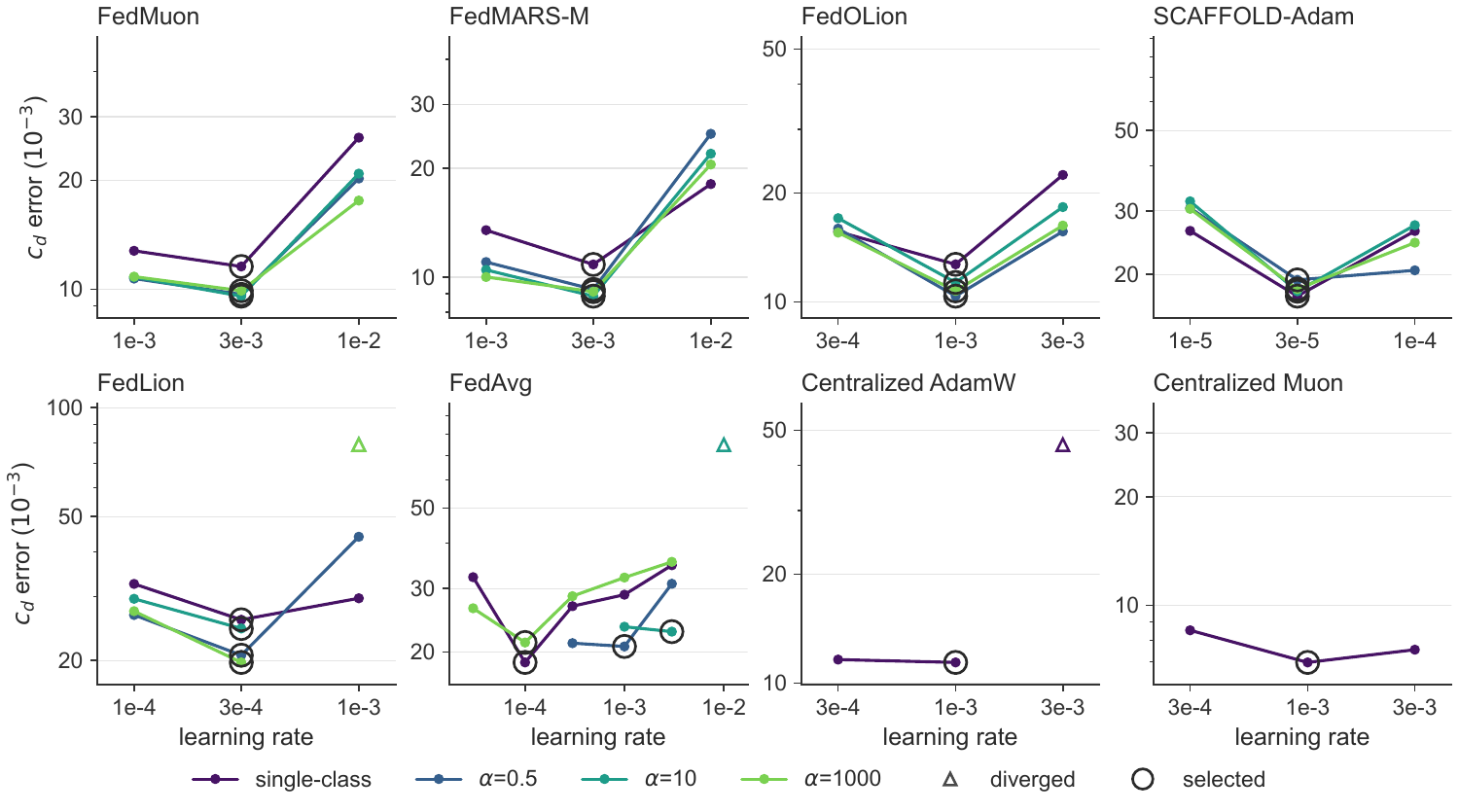}
\caption{Stage-2 learning-rate walks: \cd\ error after 30 rounds at one local epoch, one line per partition. Circles mark the selected rate, triangles a diverged run.}\label{fig:ladders}
\end{figure}

\begin{figure}[h]
\centering
\includegraphics[width=\textwidth]{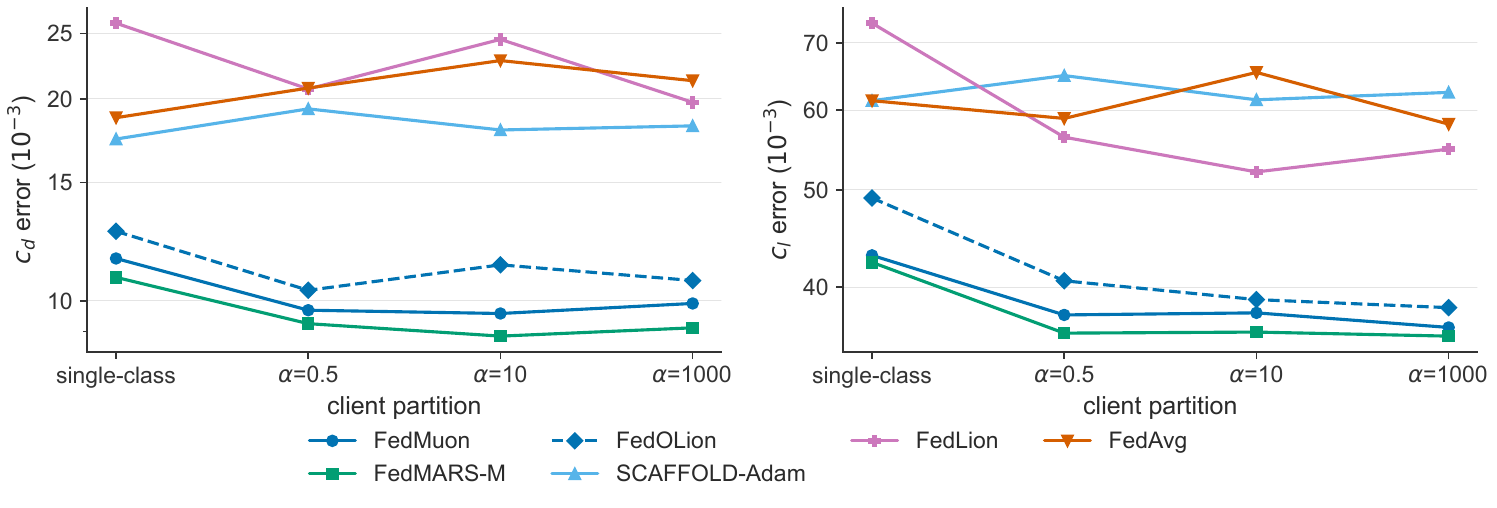}
\caption{The comparison of Figure~\ref{fig:heterogeneity} after 30 rounds, each method at its selected rate.}\label{fig:heterogeneity30}
\end{figure}

\section{Schedule ablation}\label{app:schedule}

Table~\ref{tab:schedule} repeats the local-epoch search of Table~\ref{tab:epochs} for the non-spectral methods with a constant learning rate and with the cosine schedule used in the rest of the paper. Each schedule uses its own stage-2 learning rate at one local epoch. With the learning rate re-tuned at the number of local epochs selected under each schedule, the best 30-round drag errors of FedAvg, SCAFFOLD-Adam and FedLion are 10.7, 11.9 and 12.2 with a constant learning rate and 9.2, 9.6 and 9.5 with the cosine schedule, a reduction by 14\%, 20\% and 22\%.

\begin{table}[h]{0.8\columnwidth}
\centering
\caption{Constant learning rate against the cosine schedule for the non-spectral methods: \cd\ / \cl\ error after 30 rounds on the single-class partition, units of $10^{-3}$. The constant-rate runs used A100 GPUs and the cosine runs A6000 GPUs.}\label{tab:schedule}
\footnotesize
\setlength{\tabcolsep}{5pt}
\begin{tabular*}{\linewidth}{@{\extracolsep{\fill}}llccccc@{}}
\toprule
Method & Schedule & $E = 1$ & 2 & 4 & 8 & 16 \\
\midrule
FedAvg & constant & 21.5 / 60 & 20.1 / 58 & 18.4 / 57 & 14.0 / 65 & 13.6 / 54 \\
 & cosine & 18.7 / 61 & 19.1 / 57 & 18.3 / 56 & 14.4 / 66 & 11.6 / 57 \\
SCAFFOLD-Adam & constant & 22.8 / 56 & 22.0 / 59 & 14.9 / 57 & 11.9 / 46 & 14.4 / 45 \\
 & cosine & 17.4 / 61 & 19.0 / 56 & 13.9 / 59 & 11.7 / 56 & 10.7 / 46 \\
FedLion & constant & 31.6 / 85 & 27.0 / 69 & 19.0 / 58 & 15.3 / 56 & 12.2 / 50 \\
 & cosine & 25.9 / 73 & 22.4 / 70 & 16.3 / 61 & 13.7 / 53 & 10.3 / 42 \\
\bottomrule
\end{tabular*}
\end{table}

\section{Hardware cross-check}\label{app:a6000}

The non-spectral methods were tuned on RTX A6000 GPUs and trained for 100 rounds on A100 GPUs, and we repeated their 100-round runs on A6000 GPUs. We also repeated the seed-42 runs of FedMuon and FedMARS-M at their tuned numbers of local epochs on A6000 GPUs. A run is reproducible bit for bit on A100 GPUs of either memory size but not between A100 and A6000 GPUs, so these repeats show how much a single run of the same configuration varies. Table~\ref{tab:a6000} lists the results. The drag errors of the two spectral runs change by less than 3\% ($-0.6$\% for FedMARS-M and $-2.3$\% for FedMuon), which is close to their variation across seeds. Over the 21 runs of the non-spectral methods and centralized AdamW that finished normally on both GPU types, the drag errors differ by 8\% at the median and by up to 37\%, for FedAvg at $\alpha = 10$. SCAFFOLD-Adam at its tuned number of local epochs fails on A6000 GPUs for every number of participating clients. Its loss becomes NaN with one, two and three clients, and with all six clients the training loss remains normal while the drag error of the global model ends at 294.

\begin{table}[h]{0.8\columnwidth}
\centering
\caption{The same configurations on A100 GPUs (the results of the paper) and on RTX A6000 GPUs after 100 rounds: \cd\ error / \cl\ error in units of $10^{-3}$, with $K$ clients per round.}\label{tab:a6000}
\footnotesize
\begin{tabular*}{\linewidth}{@{\extracolsep{\fill}}lllcll@{}}
\toprule
Partition & Method & $E$ & $K$ & A100 & RTX A6000 \\
\midrule
single-class   & FedMARS-M & 2 & 6 & 6.73 / 26.0 & 6.69 / 26.3 \\
               & FedMuon & 4 & 6 & 6.93 / 26.4 & 6.77 / 26.0 \\
               & FedAvg & 1 & 6 & 17.0 / 57.8 & 15.9 / 59.2 \\
               & FedAvg & 16 & 6 & 9.26 / 37.1 & 10.0 / 38.3 \\
               & FedAvg & 16 & 3 & 9.80 / 34.8 & 9.53 / 36.3 \\
               & FedAvg & 16 & 2 & 10.3 / 33.3 & 8.52 / 32.5 \\
               & FedAvg & 16 & 1 & 7.59 / 36.1 & 10.3 / 43.8 \\
               & SCAFFOLD-Adam & 1 & 6 & 13.7 / 51.1 & 13.7 / 50.6 \\
               & SCAFFOLD-Adam & 16 & 6 & 7.94 / 27.0 & 294 / 406 \\
               & SCAFFOLD-Adam & 16 & 3 & 8.01 / 28.7 & NaN at round 29 \\
               & SCAFFOLD-Adam & 16 & 2 & NaN at round 37 & NaN at round 40 \\
               & SCAFFOLD-Adam & 16 & 1 & NaN at round 36 & NaN at round 29 \\
               & FedLion & 1 & 6 & 16.4 / 51.2 & 15.2 / 55.6 \\
               & FedLion & 16 & 6 & 8.94 / 33.3 & 8.78 / 31.2 \\
               & FedLion & 16 & 3 & 9.27 / 34.6 & 9.99 / 31.8 \\
               & FedLion & 16 & 2 & 8.32 / 30.9 & 10.7 / 30.7 \\
               & FedLion & 16 & 1 & 12.6 / 44.3 & 11.1 / 35.0 \\
               & Centralized AdamW & -- & -- & 7.54 / 27.9 & 7.25 / 27.7 \\
\midrule
$\alpha = 0.5$ & FedAvg & 1 & 6 & 17.5 / 46.8 & 16.4 / 48.4 \\
               & SCAFFOLD-Adam & 1 & 6 & 14.2 / 45.9 & 14.1 / 45.5 \\
               & FedLion & 1 & 6 & 11.2 / 41.0 & 12.7 / 40.1 \\
\midrule
$\alpha = 10$  & FedAvg & 1 & 6 & 32.6 / 54.6 & 20.7 / 53.4 \\
               & SCAFFOLD-Adam & 1 & 6 & 14.4 / 45.3 & 14.0 / 45.0 \\
               & FedLion & 1 & 6 & 12.8 / 43.7 & 14.4 / 47.5 \\
\midrule
$\alpha = 1000$ & FedAvg & 1 & 6 & 16.7 / 50.8 & 16.6 / 49.8 \\
               & SCAFFOLD-Adam & 1 & 6 & 14.3 / 45.5 & 14.1 / 46.0 \\
               & FedLion & 1 & 6 & 13.2 / 45.2 & 11.9 / 44.7 \\
\bottomrule
\end{tabular*}
\end{table}

\section{Compute}\label{app:compute}

Counting training rounds only, the experiments took about 3\,950 GPU hours: 1\,820 on NVIDIA A100 80 GB GPUs, 1\,380 on RTX A6000 48 GB GPUs and 750 on A100 40 GB GPUs, with up to four GPUs of each type in use at a time. Preliminary runs with earlier versions of the protocol took another 850 GPU hours. Seed 42 of the spectral methods and of centralized Muon, all stage-1 searches and the constant-rate runs used the A100 80 GB GPUs, and seeds 43 and 44 used the A100 40 GB GPUs. Stages 2 and 3 of the non-spectral methods ran on A6000 GPUs and their 100-round runs on A100 40 GB GPUs. The two A100 types produce identical runs. Table~\ref{tab:compute} gives the time per round.

\begin{table}[h]
\centering
\caption{Wall-clock time per round of one run on one GPU. Times grow linearly with $E$ and with the number of clients per round.}\label{tab:compute}
\footnotesize
\begin{tabular*}{\linewidth}{@{\extracolsep{\fill}}lllc@{}}
\toprule
GPU & Runs & Time per round & Peak memory \\
\midrule
A100 40 GB & non-spectral, $E = 1$ & 130 s & 31.4 GB \\
A100 40 GB & non-spectral, $E = 16$ & 2\,060 s (2.4 days for 100 rounds) & 32.0 GB \\
A100 40 GB & non-spectral, $E = 16$, $K = 3$ & 1\,050 s & 32.0 GB \\
RTX A6000 48 GB & non-spectral, $E = 1$ & 233 s & 31.4 GB \\
RTX A6000 48 GB & non-spectral, $E = 16$ & 3\,820 to 4\,230 s & 31.4 GB \\
A100 40 GB & spectral, $E = 1$ & 140 s & 31.4 GB \\
A100 40 GB & FedMARS-M, $E = 2$ & 278 s & 31.4 GB \\
A100 40 GB & FedMuon, $E = 4$ & 553 s & 32.0 GB \\
A100 40 GB & FedOLion, $E = 8$ & 1\,108 s & 32.0 GB \\
\bottomrule
\end{tabular*}
\end{table}

\end{appendixpart}


\begin{thebibliography}{55}
\providecommand{\natexlab}[1]{#1}
\providecommand{\url}[1]{\texttt{#1}}
\expandafter\ifx\csname urlstyle\endcsname\relax
  \providecommand{\doi}[1]{doi: #1}\else
  \providecommand{\doi}{doi: \begingroup \urlstyle{rm}\Url}\fi

\bibitem[Ahn et~al.(2025)Ahn, Xu, Abreu, Fan, Magakyan, Sharma, Zhan, and Langford]{ahn2025dion}
Kwangjun Ahn, Byron Xu, Natalie Abreu, Ying Fan, Gagik Magakyan, Pratyusha Sharma, Zheng Zhan, and John Langford.
\newblock {D}ion: {D}istributed {O}rthonormalized {U}pdates.
\newblock \emph{arXiv preprint arXiv:2504.05295}, 2025.
\newblock URL \url{https://arxiv.org/abs/2504.05295}.

\bibitem[Bernstein and Newhouse(2024)]{bernstein2024anthology}
Jeremy Bernstein and Laker Newhouse.
\newblock {O}ld {O}ptimizer, {N}ew {N}orm: {A}n {A}nthology.
\newblock \emph{arXiv preprint arXiv:2409.20325}, 2024.
\newblock URL \url{https://arxiv.org/abs/2409.20325}.

\bibitem[Bernstein and Newhouse(2025)]{bernstein2024duality}
Jeremy Bernstein and Laker Newhouse.
\newblock {M}odular {D}uality in {D}eep {L}earning.
\newblock In \emph{Proceedings of the 42nd International Conference on Machine Learning}, 2025.
\newblock URL \url{https://arxiv.org/abs/2410.21265}.
\newblock arXiv:2410.21265.

\bibitem[Chen et~al.(2025)Chen, Zhou, Li, Wang, Gao, Shi, Zhang, and Li]{chen2025omniarch}
Tianyu Chen, Haoyi Zhou, Ying Li, Hao Wang, Chonghan Gao, Rongye Shi, Shanghang Zhang, and Jianxin Li.
\newblock {OmniArch}: {B}uilding {F}oundation {M}odel for {S}cientific {C}omputing.
\newblock In \emph{Proceedings of the 42nd International Conference on Machine Learning}, 2025.
\newblock URL \url{https://arxiv.org/abs/2402.16014}.
\newblock arXiv:2402.16014.

\bibitem[Chen et~al.(2023)Chen, Liang, Huang, Real, Wang, Pham, Dong, Luong, Hsieh, Lu, and Le]{chen2023lion}
Xiangning Chen, Chen Liang, Da~Huang, Esteban Real, Kaiyuan Wang, Hieu Pham, Xuanyi Dong, Thang Luong, Cho-Jui Hsieh, Yifeng Lu, and Quoc~V. Le.
\newblock {S}ymbolic {D}iscovery of {O}ptimization {A}lgorithms.
\newblock In \emph{Advances in Neural Information Processing Systems}, volume~36, 2023.
\newblock URL \url{https://arxiv.org/abs/2302.06675}.

\bibitem[Dayan et~al.(2021)Dayan, Roth, Zhong, et~al.]{dayan2021covid}
Ittai Dayan, Holger~R. Roth, Aoxiao Zhong, et~al.
\newblock {F}ederated {L}earning for {P}redicting {C}linical {O}utcomes in {P}atients with {COVID}-19.
\newblock \emph{Nature Medicine}, 27\penalty0 (10):\penalty0 1735--1743, 2021.
\newblock URL \url{https://doi.org/10.1038/s41591-021-01506-3}.

\bibitem[Elrefaie et~al.(2024)Elrefaie, Morar, Dai, and Ahmed]{elrefaie2024drivaernetpp}
Mohamed Elrefaie, Florin Morar, Angela Dai, and Faez Ahmed.
\newblock {DrivAerNet++}: {A} {L}arge-{S}cale {M}ultimodal {C}ar {D}ataset with {C}omputational {F}luid {D}ynamics {S}imulations and {D}eep {L}earning {B}enchmarks.
\newblock In \emph{Advances in Neural Information Processing Systems, Datasets and Benchmarks Track}, 2024.
\newblock URL \url{https://arxiv.org/abs/2406.09624}.

\bibitem[Gorbunov et~al.(2021)Gorbunov, Hanzely, and Richt{\'a}rik]{gorbunov2021local}
Eduard Gorbunov, Filip Hanzely, and Peter Richt{\'a}rik.
\newblock {L}ocal {SGD}: {U}nified {T}heory and {N}ew {E}fficient {M}ethods.
\newblock In \emph{Proceedings of the 24th International Conference on Artificial Intelligence and Statistics}, pages 3556--3564, 2021.
\newblock URL \url{https://proceedings.mlr.press/v130/gorbunov21a.html}.

\bibitem[Goyal et~al.(2017)Goyal, Doll{\'a}r, Girshick, Noordhuis, Wesolowski, Kyrola, Tulloch, Jia, and He]{goyal2017accurate}
Priya Goyal, Piotr Doll{\'a}r, Ross Girshick, Pieter Noordhuis, Lukasz Wesolowski, Aapo Kyrola, Andrew Tulloch, Yangqing Jia, and Kaiming He.
\newblock {A}ccurate, {L}arge {M}inibatch {SGD}: {T}raining {ImageNet} in 1 {H}our.
\newblock \emph{arXiv preprint arXiv:1706.02677}, 2017.
\newblock URL \url{https://arxiv.org/abs/1706.02677}.

\bibitem[Hard et~al.(2018)Hard, Rao, Mathews, et~al.]{hard2018keyboard}
Andrew Hard, Kanishka Rao, Rajiv Mathews, et~al.
\newblock {F}ederated {L}earning for {M}obile {K}eyboard {P}rediction.
\newblock \emph{arXiv preprint arXiv:1811.03604}, 2018.
\newblock URL \url{https://arxiv.org/abs/1811.03604}.

\bibitem[Hoffmann et~al.(2022)Hoffmann, Borgeaud, Mensch, Buchatskaya, Cai, Rutherford, et~al.]{hoffmann2022chinchilla}
Jordan Hoffmann, Sebastian Borgeaud, Arthur Mensch, Elena Buchatskaya, Trevor Cai, Eliza Rutherford, et~al.
\newblock {T}raining {C}ompute-{O}ptimal {L}arge {L}anguage {M}odels.
\newblock In \emph{Advances in Neural Information Processing Systems}, volume~35, 2022.
\newblock URL \url{https://arxiv.org/abs/2203.15556}.

\bibitem[Hsieh et~al.(2020)Hsieh, Phanishayee, Mutlu, and Gibbons]{hsieh2020quagmire}
Kevin Hsieh, Amar Phanishayee, Onur Mutlu, and Phillip~B. Gibbons.
\newblock {T}he {N}on-{IID} {D}ata {Q}uagmire of {D}ecentralized {M}achine {L}earning.
\newblock In \emph{Proceedings of the 37th International Conference on Machine Learning}, 2020.
\newblock URL \url{https://arxiv.org/abs/1910.00189}.

\bibitem[Hsu et~al.(2019)Hsu, Qi, and Brown]{hsu2019dirichlet}
Tzu-Ming~Harry Hsu, Hang Qi, and Matthew Brown.
\newblock {M}easuring the {E}ffects of {N}on-{I}dentical {D}ata {D}istribution for {F}ederated {V}isual {C}lassification.
\newblock \emph{arXiv preprint arXiv:1909.06335}, 2019.
\newblock URL \url{https://arxiv.org/abs/1909.06335}.

\bibitem[Jordan et~al.(2024)Jordan, Jin, Boza, You, Cesista, Newhouse, and Bernstein]{jordan2024muon}
Keller Jordan, Yuchen Jin, Vlado Boza, Jiacheng You, Franz Cesista, Laker Newhouse, and Jeremy Bernstein.
\newblock {M}uon: {A}n {O}ptimizer for {H}idden {L}ayers in {N}eural {N}etworks.
\newblock \url{https://kellerjordan.github.io/posts/muon/}, 2024.

\bibitem[Kairouz et~al.(2021)Kairouz, McMahan, et~al.]{kairouz2021advances}
Peter Kairouz, H.~Brendan McMahan, et~al.
\newblock {A}dvances and {O}pen {P}roblems in {F}ederated {L}earning.
\newblock \emph{Foundations and Trends in Machine Learning}, 14\penalty0 (1--2):\penalty0 1--210, 2021.
\newblock URL \url{https://doi.org/10.1561/2200000083}.

\bibitem[Karimireddy et~al.(2020)Karimireddy, Kale, Mohri, Reddi, Stich, and Suresh]{karimireddy2020scaffold}
Sai~Praneeth Karimireddy, Satyen Kale, Mehryar Mohri, Sashank~J. Reddi, Sebastian Stich, and Ananda~Theertha Suresh.
\newblock {SCAFFOLD}: {S}tochastic {C}ontrolled {A}veraging for {F}ederated {L}earning.
\newblock In \emph{Proceedings of the 37th International Conference on Machine Learning}, 2020.
\newblock URL \url{https://arxiv.org/abs/1910.06378}.

\bibitem[Khaled et~al.(2020)Khaled, Mishchenko, and Richt{\'a}rik]{khaled2020tighter}
Ahmed Khaled, Konstantin Mishchenko, and Peter Richt{\'a}rik.
\newblock {T}ighter {T}heory for {L}ocal {SGD} on {I}dentical and {H}eterogeneous {D}ata.
\newblock In \emph{Proceedings of the 23rd International Conference on Artificial Intelligence and Statistics}, pages 4519--4529, 2020.
\newblock URL \url{https://proceedings.mlr.press/v108/bayoumi20a.html}.

\bibitem[Kone{\v{c}}n{\'y} et~al.(2016{\natexlab{a}})Kone{\v{c}}n{\'y}, McMahan, Ramage, and Richt{\'a}rik]{konecny2016federatedopt}
Jakub Kone{\v{c}}n{\'y}, H.~Brendan McMahan, Daniel Ramage, and Peter Richt{\'a}rik.
\newblock {F}ederated {O}ptimization: {D}istributed {M}achine {L}earning for {O}n-{D}evice {I}ntelligence.
\newblock \emph{arXiv preprint arXiv:1610.02527}, 2016{\natexlab{a}}.
\newblock URL \url{https://arxiv.org/abs/1610.02527}.

\bibitem[Kone{\v{c}}n{\'y} et~al.(2016{\natexlab{b}})Kone{\v{c}}n{\'y}, McMahan, Yu, Richt{\'a}rik, Suresh, and Bacon]{konecny2016federated}
Jakub Kone{\v{c}}n{\'y}, H.~Brendan McMahan, Felix~X. Yu, Peter Richt{\'a}rik, Ananda~Theertha Suresh, and Dave Bacon.
\newblock {F}ederated {L}earning: {S}trategies for {I}mproving {C}ommunication {E}fficiency.
\newblock \emph{arXiv preprint arXiv:1610.05492}, 2016{\natexlab{b}}.
\newblock URL \url{https://arxiv.org/abs/1610.05492}.

\bibitem[Kovalev(2025)]{kovalev2025trust}
Dmitry Kovalev.
\newblock {U}nderstanding {G}radient {O}rthogonalization for {D}eep {L}earning via {N}on-{E}uclidean {T}rust-{R}egion {O}ptimization.
\newblock \emph{arXiv preprint arXiv:2503.12645}, 2025.
\newblock URL \url{https://arxiv.org/abs/2503.12645}.

\bibitem[Krizhevsky(2009)]{krizhevsky2009cifar}
Alex Krizhevsky.
\newblock {L}earning {M}ultiple {L}ayers of {F}eatures from {T}iny {I}mages.
\newblock Technical report, University of Toronto, 2009.
\newblock URL \url{https://www.cs.toronto.edu/~kriz/learning-features-2009-TR.pdf}.

\bibitem[LeCun et~al.(1998)LeCun, Bottou, Bengio, and Haffner]{lecun1998mnist}
Yann LeCun, L{\'e}on Bottou, Yoshua Bengio, and Patrick Haffner.
\newblock {G}radient-{B}ased {L}earning {A}pplied to {D}ocument {R}ecognition.
\newblock \emph{Proceedings of the IEEE}, 86\penalty0 (11):\penalty0 2278--2324, 1998.
\newblock URL \url{https://doi.org/10.1109/5.726791}.

\bibitem[Li and Hong(2025)]{li2025note}
Jiaxiang Li and Mingyi Hong.
\newblock {A} {N}ote on the {C}onvergence of {M}uon.
\newblock \emph{arXiv preprint arXiv:2502.02900}, 2025.
\newblock URL \url{https://arxiv.org/abs/2502.02900}.

\bibitem[Li et~al.(2022)Li, Diao, Chen, and He]{li2022silos}
Qinbin Li, Yiqun Diao, Quan Chen, and Bingsheng He.
\newblock {F}ederated {L}earning on {N}on-{IID} {D}ata {S}ilos: {A}n {E}xperimental {S}tudy.
\newblock In \emph{IEEE 38th International Conference on Data Engineering}, pages 965--978, 2022.
\newblock URL \url{https://doi.org/10.1109/ICDE53745.2022.00077}.

\bibitem[Li et~al.(2020{\natexlab{a}})Li, Sahu, Talwalkar, and Smith]{li2020challenges}
Tian Li, Anit~Kumar Sahu, Ameet Talwalkar, and Virginia Smith.
\newblock {F}ederated {L}earning: {C}hallenges, {M}ethods, and {F}uture {D}irections.
\newblock \emph{IEEE Signal Processing Magazine}, 37\penalty0 (3):\penalty0 50--60, 2020{\natexlab{a}}.
\newblock URL \url{https://doi.org/10.1109/MSP.2020.2975749}.

\bibitem[Li et~al.(2020{\natexlab{b}})Li, Sahu, Zaheer, Sanjabi, Talwalkar, and Smith]{li2020fedprox}
Tian Li, Anit~Kumar Sahu, Manzil Zaheer, Maziar Sanjabi, Ameet Talwalkar, and Virginia Smith.
\newblock {F}ederated {O}ptimization in {H}eterogeneous {N}etworks.
\newblock In \emph{Proceedings of Machine Learning and Systems}, 2020{\natexlab{b}}.
\newblock URL \url{https://arxiv.org/abs/1812.06127}.

\bibitem[Li et~al.(2021)Li, Kovachki, Azizzadenesheli, Liu, Bhattacharya, Stuart, and Anandkumar]{li2021fno}
Zongyi Li, Nikola Kovachki, Kamyar Azizzadenesheli, Burigede Liu, Kaushik Bhattacharya, Andrew Stuart, and Anima Anandkumar.
\newblock {F}ourier {N}eural {O}perator for {P}arametric {P}artial {D}ifferential {E}quations.
\newblock In \emph{International Conference on Learning Representations}, 2021.
\newblock URL \url{https://arxiv.org/abs/2010.08895}.

\bibitem[Liu et~al.(2025{\natexlab{a}})Liu, Su, Yao, Jiang, Lai, Du, et~al.]{liu2025muonscalable}
Jingyuan Liu, Jianlin Su, Xingcheng Yao, Zhejun Jiang, Guokun Lai, Yulun Du, et~al.
\newblock {M}uon {I}s {S}calable for {LLM} {T}raining.
\newblock \emph{arXiv preprint arXiv:2502.16982}, 2025{\natexlab{a}}.
\newblock URL \url{https://arxiv.org/abs/2502.16982}.

\bibitem[Liu et~al.(2025{\natexlab{b}})Liu, Shang, Zhou, Liu, Liu, and Liu]{liu2025fedmuon}
Junkang Liu, Fanhua Shang, Junchao Zhou, Hongying Liu, Yuanyuan Liu, and Jin Liu.
\newblock {FedMuon}: {A}ccelerating {F}ederated {L}earning with {M}atrix {O}rthogonalization.
\newblock \emph{arXiv preprint arXiv:2510.27403}, 2025{\natexlab{b}}.
\newblock URL \url{https://arxiv.org/abs/2510.27403}.

\bibitem[Loshchilov and Hutter(2017)]{loshchilov2017sgdr}
Ilya Loshchilov and Frank Hutter.
\newblock {SGDR}: {S}tochastic {G}radient {D}escent with {W}arm {R}estarts.
\newblock In \emph{International Conference on Learning Representations}, 2017.
\newblock URL \url{https://arxiv.org/abs/1608.03983}.

\bibitem[Loshchilov and Hutter(2019)]{loshchilov2019adamw}
Ilya Loshchilov and Frank Hutter.
\newblock {D}ecoupled {W}eight {D}ecay {R}egularization.
\newblock In \emph{International Conference on Learning Representations}, 2019.
\newblock URL \url{https://arxiv.org/abs/1711.05101}.

\bibitem[Lu et~al.(2021)Lu, Jin, Pang, Zhang, and Karniadakis]{lu2021deeponet}
Lu~Lu, Pengzhan Jin, Guofei Pang, Zhongqiang Zhang, and George~Em Karniadakis.
\newblock {L}earning {N}onlinear {O}perators via {DeepONet} {B}ased on the {U}niversal {A}pproximation {T}heorem of {O}perators.
\newblock \emph{Nature Machine Intelligence}, 3:\penalty0 218--229, 2021.
\newblock URL \url{https://doi.org/10.1038/s42256-021-00302-5}.

\bibitem[McMahan et~al.(2017)McMahan, Moore, Ramage, Hampson, and Ag{\"u}era~y Arcas]{mcmahan2017fedavg}
H.~Brendan McMahan, Eider Moore, Daniel Ramage, Seth Hampson, and Blaise Ag{\"u}era~y Arcas.
\newblock {C}ommunication-{E}fficient {L}earning of {D}eep {N}etworks from {D}ecentralized {D}ata.
\newblock In \emph{Proceedings of the 20th International Conference on Artificial Intelligence and Statistics}, 2017.
\newblock URL \url{https://arxiv.org/abs/1602.05629}.

\bibitem[Mishchenko et~al.(2022)Mishchenko, Malinovsky, Stich, and Richt{\'a}rik]{mishchenko2022proxskip}
Konstantin Mishchenko, Grigory Malinovsky, Sebastian Stich, and Peter Richt{\'a}rik.
\newblock {ProxSkip}: {Y}es! {L}ocal {G}radient {S}teps {P}rovably {L}ead to {C}ommunication {A}cceleration! {F}inally!
\newblock In \emph{Proceedings of the 39th International Conference on Machine Learning}, pages 15750--15769, 2022.
\newblock URL \url{https://proceedings.mlr.press/v162/mishchenko22b.html}.

\bibitem[Pethick et~al.(2025)Pethick, Xie, Antonakopoulos, Zhu, Silveti-Falls, and Cevher]{pethick2025lmo}
Thomas Pethick, Wanyun Xie, Kimon Antonakopoulos, Zhenyu Zhu, Antonio Silveti-Falls, and Volkan Cevher.
\newblock {T}raining {D}eep {L}earning {M}odels with {N}orm-{C}onstrained {LMOs}.
\newblock In \emph{Proceedings of the 42nd International Conference on Machine Learning}, 2025.
\newblock URL \url{https://arxiv.org/abs/2502.07529}.

\bibitem[Pfaff et~al.(2021)Pfaff, Fortunato, Sanchez-Gonzalez, and Battaglia]{pfaff2021meshgraphnets}
Tobias Pfaff, Meire Fortunato, Alvaro Sanchez-Gonzalez, and Peter~W. Battaglia.
\newblock {L}earning {M}esh-{B}ased {S}imulation with {G}raph {N}etworks.
\newblock In \emph{International Conference on Learning Representations}, 2021.
\newblock URL \url{https://arxiv.org/abs/2010.03409}.

\bibitem[Reddi et~al.(2021)Reddi, Charles, Zaheer, Garrett, Rush, Kone{\v{c}}n{\'y}, Kumar, and McMahan]{reddi2021fedopt}
Sashank Reddi, Zachary Charles, Manzil Zaheer, Zachary Garrett, Keith Rush, Jakub Kone{\v{c}}n{\'y}, Sanjiv Kumar, and H.~Brendan McMahan.
\newblock {A}daptive {F}ederated {O}ptimization.
\newblock In \emph{International Conference on Learning Representations}, 2021.
\newblock URL \url{https://arxiv.org/abs/2003.00295}.

\bibitem[Riabinin et~al.(2025)Riabinin, Shulgin, Gruntkowska, and Richt{\'a}rik]{riabinin2025gluon}
Artem Riabinin, Egor Shulgin, Kaja Gruntkowska, and Peter Richt{\'a}rik.
\newblock {G}luon: {M}aking {M}uon \& {S}cion {G}reat {A}gain! ({B}ridging {T}heory and {P}ractice of {LMO}-{B}ased {O}ptimizers for {LLMs}).
\newblock \emph{arXiv preprint arXiv:2505.13416}, 2025.
\newblock URL \url{https://arxiv.org/abs/2505.13416}.

\bibitem[Riabinin et~al.(2026)Riabinin, Veprikov, Bolatov, Tak{\'a}{\v{c}}, and Beznosikov]{riabinin2026warmup}
Artem Riabinin, Andrey Veprikov, Arman Bolatov, Martin Tak{\'a}{\v{c}}, and Aleksandr Beznosikov.
\newblock {W}here {D}oes {W}arm-{U}p {C}ome {F}rom? {A}daptive {S}cheduling for {N}orm-{C}onstrained {O}ptimizers.
\newblock \emph{arXiv preprint arXiv:2602.05813}, 2026.
\newblock URL \url{https://arxiv.org/abs/2602.05813}.

\bibitem[Rieke et~al.(2020)Rieke, Hancox, Li, et~al.]{rieke2020health}
Nicola Rieke, Jonny Hancox, Wenqi Li, et~al.
\newblock {T}he {F}uture of {D}igital {H}ealth with {F}ederated {L}earning.
\newblock \emph{npj Digital Medicine}, 3:\penalty0 119, 2020.
\newblock URL \url{https://doi.org/10.1038/s41746-020-00323-1}.

\bibitem[Sheller et~al.(2020)Sheller, Edwards, Reina, et~al.]{sheller2020medicine}
Micah~J. Sheller, Brandon Edwards, G.~Anthony Reina, et~al.
\newblock {F}ederated {L}earning in {M}edicine: {F}acilitating {M}ulti-{I}nstitutional {C}ollaborations {W}ithout {S}haring {P}atient {D}ata.
\newblock \emph{Scientific Reports}, 10:\penalty0 12598, 2020.
\newblock URL \url{https://doi.org/10.1038/s41598-020-69250-1}.

\bibitem[Shen et~al.(2026)Shen, Huang, Huang, Shen, and Zhang]{shen2025convergence}
Wei Shen, Ruichuan Huang, Minhui Huang, Cong Shen, and Jiawei Zhang.
\newblock {O}n the {C}onvergence {A}nalysis of {M}uon.
\newblock \emph{Transactions on Machine Learning Research}, 2026.
\newblock URL \url{https://arxiv.org/abs/2505.23737}.
\newblock arXiv:2505.23737.

\bibitem[Stich(2019)]{stich2019local}
Sebastian~U. Stich.
\newblock {L}ocal {SGD} {C}onverges {F}ast and {C}ommunicates {L}ittle.
\newblock In \emph{International Conference on Learning Representations}, 2019.
\newblock URL \url{https://arxiv.org/abs/1805.09767}.

\bibitem[Takezawa et~al.(2026)Takezawa, Koloskova, Jiang, and Stich]{takezawa2025fedmuon}
Yuki Takezawa, Anastasia Koloskova, Xiaowen Jiang, and Sebastian~U. Stich.
\newblock {FedMuon}: {F}ederated {L}earning with {B}ias-{C}orrected {LMO}-{B}ased {O}ptimization.
\newblock In \emph{International Conference on Learning Representations}, 2026.
\newblock URL \url{https://arxiv.org/abs/2509.26337}.
\newblock arXiv:2509.26337.

\bibitem[Tang and Chang(2024)]{tang2024fedlion}
Zhiwei Tang and Tsung-Hui Chang.
\newblock {FedLion}: {F}aster {A}daptive {F}ederated {O}ptimization with {F}ewer {C}ommunication.
\newblock In \emph{IEEE International Conference on Acoustics, Speech and Signal Processing (ICASSP)}, pages 13316--13320, 2024.
\newblock URL \url{https://doi.org/10.1109/ICASSP48485.2024.10447045}.

\bibitem[Th{\'e}rien et~al.(2026)Th{\'e}rien, Huang, Defazio, Rish, and Belilovsky]{therien2025muloco}
Benjamin Th{\'e}rien, Xiaolong Huang, Aaron Defazio, Irina Rish, and Eugene Belilovsky.
\newblock {MuLoCo}: {M}uon {I}s a {P}ractical {I}nner {O}ptimizer for {DiLoCo}.
\newblock In \emph{Proceedings of the 43rd International Conference on Machine Learning}, 2026.
\newblock URL \url{https://arxiv.org/abs/2505.23725}.
\newblock arXiv:2505.23725.

\bibitem[Veprikov et~al.(2025)Veprikov, Bolatov, Bogdanov, Horv{\'a}th, Beznosikov, Tak{\'a}{\v{c}}, and Hanzely]{veprikov2025preconditioned}
Andrey Veprikov, Arman Bolatov, Aleksandr Bogdanov, Samuel Horv{\'a}th, Aleksandr Beznosikov, Martin Tak{\'a}{\v{c}}, and Slavom{\'i}r Hanzely.
\newblock {P}reconditioned {N}orms: {A} {U}nified {F}ramework for {S}teepest {D}escent, {Q}uasi-{N}ewton and {A}daptive {M}ethods.
\newblock \emph{arXiv preprint arXiv:2510.10777}, 2025.
\newblock URL \url{https://arxiv.org/abs/2510.10777}.

\bibitem[Wang and Joshi(2021)]{wang2018cooperative}
Jianyu Wang and Gauri Joshi.
\newblock {C}ooperative {SGD}: {A} {U}nified {F}ramework for the {D}esign and {A}nalysis of {L}ocal-{U}pdate {SGD} {A}lgorithms.
\newblock \emph{Journal of Machine Learning Research}, 22\penalty0 (213):\penalty0 1--50, 2021.
\newblock URL \url{https://jmlr.org/papers/v22/20-147.html}.

\bibitem[Wang et~al.(2021)Wang, Charles, Xu, et~al.]{wang2021fieldguide}
Jianyu Wang, Zachary Charles, Zheng Xu, et~al.
\newblock {A} {F}ield {G}uide to {F}ederated {O}ptimization.
\newblock \emph{arXiv preprint arXiv:2107.06917}, 2021.
\newblock URL \url{https://arxiv.org/abs/2107.06917}.

\bibitem[Wang et~al.(2026)Wang, Shen, and Zhang]{wang2026olion}
Zixiao Wang, Yifei Shen, and Huishuai Zhang.
\newblock {OLion}: {A}pproaching the {H}adamard {I}deal by {I}ntersecting {S}pectral and $\ell_{\infty}$ {I}mplicit {B}iases.
\newblock In \emph{Proceedings of the 43rd International Conference on Machine Learning}, 2026.
\newblock URL \url{https://arxiv.org/abs/2602.01105}.
\newblock arXiv:2602.01105.

\bibitem[Wu et~al.(2024)Wu, Luo, Wang, Wang, and Long]{wu2024transolver}
Haixu Wu, Huakun Luo, Haowen Wang, Jianmin Wang, and Mingsheng Long.
\newblock {T}ransolver: {A} {F}ast {T}ransformer {S}olver for {PDEs} on {G}eneral {G}eometries.
\newblock In \emph{Proceedings of the 41st International Conference on Machine Learning}, pages 53681--53705, 2024.
\newblock URL \url{https://arxiv.org/abs/2402.02366}.

\bibitem[Yang et~al.(2019)Yang, Liu, Chen, and Tong]{yang2019concept}
Qiang Yang, Yang Liu, Tianjian Chen, and Yongxin Tong.
\newblock {F}ederated {M}achine {L}earning: {C}oncept and {A}pplications.
\newblock \emph{ACM Transactions on Intelligent Systems and Technology}, 10\penalty0 (2), 2019.
\newblock URL \url{https://doi.org/10.1145/3298981}.

\bibitem[Yuan et~al.(2025)Yuan, Liu, Wu, Zhou, and Gu]{yuan2024mars}
Huizhuo Yuan, Yifeng Liu, Shuang Wu, Xun Zhou, and Quanquan Gu.
\newblock {MARS}: {U}nleashing the {P}ower of {V}ariance {R}eduction for {T}raining {L}arge {M}odels.
\newblock In \emph{Proceedings of the 42nd International Conference on Machine Learning}, 2025.
\newblock URL \url{https://arxiv.org/abs/2411.10438}.

\bibitem[Zhang and Gao(2025)]{zhang2025provable}
Xinwen Zhang and Hongchang Gao.
\newblock {O}n {P}rovable {B}enefits of {M}uon in {F}ederated {L}earning.
\newblock \emph{arXiv preprint arXiv:2510.03866}, 2025.
\newblock URL \url{https://arxiv.org/abs/2510.03866}.

\bibitem[Zhao et~al.(2018)Zhao, Li, Lai, Suda, Civin, and Chandra]{zhao2018noniid}
Yue Zhao, Meng Li, Liangzhen Lai, Naveen Suda, Damon Civin, and Vikas Chandra.
\newblock {F}ederated {L}earning with {N}on-{IID} {D}ata.
\newblock \emph{arXiv preprint arXiv:1806.00582}, 2018.
\newblock URL \url{https://arxiv.org/abs/1806.00582}.

\end{thebibliography}
\end{document}